\documentclass[runningheads]{llncs}

\usepackage{author-kit/eccv}

\usepackage{author-kit/eccvabbrv}

\usepackage{graphicx}
\usepackage{booktabs}

\usepackage[accsupp]{axessibility}  

\usepackage{amsmath,amsfonts,bm}
\usepackage{url}
\usepackage{fontawesome5}
\usepackage{twemojis}
\usepackage{comment}
\usepackage{xspace}
\usepackage{graphicx}
\usepackage{booktabs}
\usepackage{mwe}
\usepackage{subcaption}
\usepackage{pdflscape}
\usepackage{adjustbox} 
\usepackage{colortbl}
\usepackage[dvipsnames]{xcolor}
\usepackage{pifont}
\usepackage{wrapfig}
\usepackage{enumitem}
\usepackage{adjustbox}
\usepackage{longtable}
\usepackage{svg}
\usepackage{makecell}
\usepackage{algorithm}
\usepackage{algpseudocode}
\usepackage{microtype}
\usepackage[framemethod=tikz]{mdframed} 
\newcommand{\cmark}{\textcolor{ForestGreen}{\ding{51}}}%
\newcommand{\xmark}{\textcolor{red}{\ding{55}}}%
\newcommand{\ourDataset}{HouseCorr3D\xspace}
\newcommand{\ours}{Morpheus\xspace}

\newcommand{\affinef}{\phi_{a}}
\newcommand{\sdff}{\phi_{sdf}}

\newcommand{\affinefscale}{\alpha}
\newcommand{\affineftransl}{\delta}

\newcommand{\vertexsymbol}{\mathrm{v}}

\def\vv{{\bm{v}}}
\def\vi{{\bm{i}}}
\def\vj{{\bm{j}}}
\newcommand{\vei}{\vertexsymbol_i}
\newcommand{\vej}{\vertexsymbol_j}

\def\vx{{\bm{x}}}

\newcommand{\instlatent}{\mathrm{l}}
\newcommand{\instlatentspace}{\mathrm{L}}
\newcommand{\query}{q}
\newcommand{\target}{t}

\newcommand{\featencl}{\psi_\instlatent}
\newcommand{\xquery}{x^{\query}}

\newcommand{\meshdefquery}{\meshdef^{\query}}
\newcommand{\meshquery}{\mesh^{\query}}
\newcommand{\xtarget}{x^{\target}}
\newcommand{\xtargetproj}{\hat{x}^{\target}}
\newcommand{\meshdeftarget}{\meshdef^{\target}}
\newcommand{\meshtarget}{\mesh^{\target}}

\newcommand{\edges}{\mathrm{E}}

\newcommand{\pose}{\pi}

\newcommand{\pointssdf}{\mathcal{P}_{sdf}}

\newcommand{\mask}{\mathrm{m}}
\newcommand{\maskpred}{\tilde{\mask}}
\newcommand{\dt}{\texttt{dt}}
\newcommand{\maskdt}{\dt(\mask)}

\newcommand{\lossmask}{\mathcal{L}_\mask}
\newcommand{\lossmaskdt}{\mathcal{L}_{\mask\mathrm{dt}}}
\newcommand{\losseik}{\mathcal{L}_{sdf}}
\newcommand{\lossrec}{\mathcal{L}_{CD}}
\newcommand{\lossregdef}{\mathcal{L}_{def}}
\newcommand{\lossregdefsm}{\mathcal{L}_{smooth}}

\newcommand{\glossmask}{\lambda_\mask}
\newcommand{\glossmaskdt}{\lambda_{\mask\mathrm{dt}}}
\newcommand{\glosseik}{\lambda_{sdf}}
\newcommand{\glossrec}{\lambda_{CD}}
\newcommand{\glossregdef}{\lambda_{def}}
\newcommand{\glossregdefsm}{\lambda_{smooth}}

\newcommand{\vnneuc}{\chi{}}  

\newcommand{\imagesymbol}{\mathrm{I}}

\newcommand{\imagesymbolquery}{\imagesymbol^{\query}}
\newcommand{\imagesymboltarget}{\imagesymbol^{\target}}

\newcommand{\verts}{\mathrm{V}}
\newcommand{\vertsdef}{\mathrm{V}_{def}}
\newcommand{\vertsdefgt}{\mathrm{V}_{gt}}

\newcommand{\mesh}{\mathrm{M}}
\newcommand{\meshdef}{\mathrm{M}_{def}}
\newcommand{\meshgt}{\mathrm{M}_{gt}}

\newcommand{\Backpack}[1][]{\includegraphics[height=10pt]{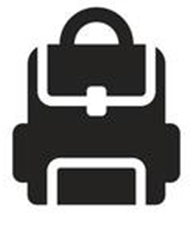}}
\definecolor{pipelinepurple}{RGB}{150,115,166}
\definecolor{pipelineorange}{RGB}{255, 128, 0}
\newcommand{\doublehline}{\hline\noalign{\vskip 0.2ex}\hline}

\usepackage[pagebackref,breaklinks,colorlinks,citecolor=eccvblue]{hyperref}

\usepackage{orcidlink}

\begin{document}

\title{\large Category-Level 3D Correspondence in Camera Space\\  via Morphable Object Priors}

\makeatletter
\def\thanks#1{\protected@xdef\@thanks{\@thanks
        \protect\footnotetext{#1}}}
\makeatother

\newcommand*\samethanks[1][\value{footnote}]{\footnotemark[#1]}

\author{
Leonhard Sommer\inst{1}\thanks{$^\star$ \ Equal contribution.}\samethanks[1]\textsuperscript{\faEnvelope[regular]}\thanks{\textsuperscript{\faEnvelope[regular]} Corresponding authors: \email{\{sommerl,jesslen\}@cs.uni-freiburg.de}.}\orcidlink{000-0003-0876-2942} \and
Artur Jesslen\inst{1}\samethanks[1]\textsuperscript{\faEnvelope[regular]}\orcidlink{0000-0002-4837-8163} \and \\
Basavaraj Sunagad\inst{2}\orcidlink{0009-0009-8618-2805} \and
Adam Kortylewski\inst{2}\orcidlink{0000-0002-9146-4403}
}

\authorrunning{Sommer et al.}
\titlerunning{HouseCorr3D: Category-Level 3D Correspondence}

\institute{University of Freiburg, Germany \and
CISPA Helmholtz Center for Information Security, Germany}

\maketitle
\begin{abstract}
Understanding 3D objects from images is fundamental to robotics and AR/VR applications.
While recent work has made progress in category-level pose estimation, current representations fail to capture the fine-grained semantics needed for reasoning about object parts, functions, and interactions.
In this work, we study \textbf{category-level 3D correspondence in camera space}---predicting, from a single image, 3D locations that remain consistent across instances within a category---and show that it can emerge without explicit correspondence supervision by learning a shared morphable object prior.
To enable research in this direction, we introduce \textbf{\ourDataset}, the first large-scale benchmark for monocular category-level 3D correspondence with 178k images across 50 household object categories, 280 unique instances, and 3D keypoint annotations directly on CAD models.
Crucially, \ourDataset provides amodal correspondence labels for occluded regions and explicit symmetry annotations, addressing key limitations of existing datasets.
We further propose \textbf{\ours}, a method that learns \textbf{morphable category-level shape priors} by disentangling canonical shape, deformation, and object pose. Through this shared canonical grounding, semantically meaningful 3D correspondences \textbf{in camera space} emerge implicitly. These emerging 3D correspondences set a new state of the art on \ourDataset, demonstrating that semantic 3D object understanding can arise without direct correspondence supervision. Data and code: \href{https://github.com/GenIntel/HouseCorr3D}{\faGithub/GenIntel/HouseCorr3D}.
\end{abstract}
    
\section{Introduction}
\begin{figure*}
    \centering
    \includegraphics[width=\linewidth]{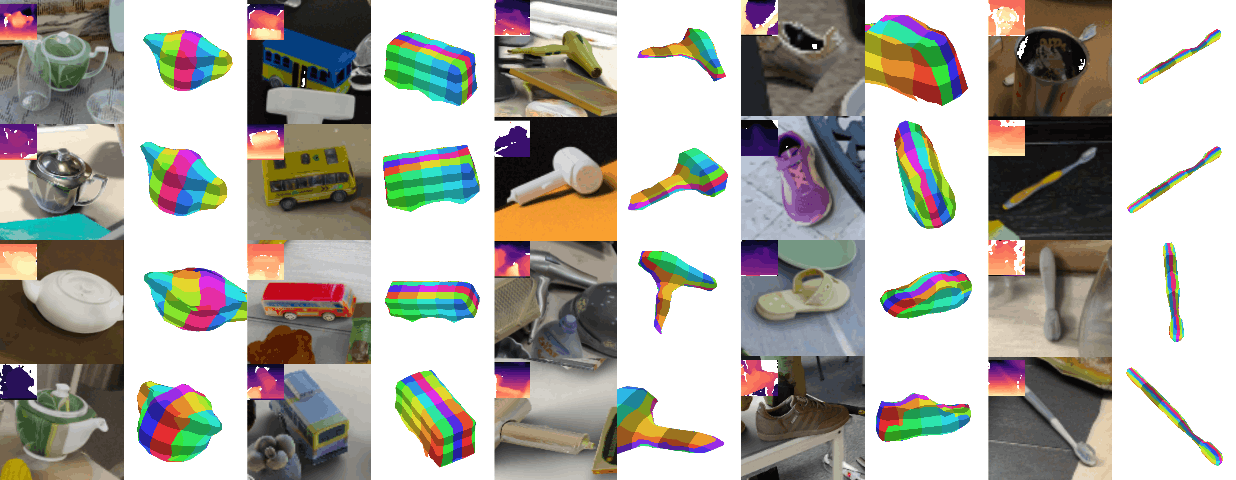}
    \caption{\textbf{Monocular Category-level 3D Correspondence.} We predict semantically consistent 3D keypoint locations across different instances of the same category from single RGB-D images. Our morphable priors enable establishing correspondences (shown with matching colors) that remain semantically aligned despite large shape variations, enabling fine-grained object understanding beyond traditional pose estimation and 2D semantic correspondence.}
    \label{fig:teaser}
    \vspace{-1em}
\end{figure*}

Understanding objects in 3D from images is a long-standing challenge in computer vision, with applications in robotics, augmented reality (AR), and virtual reality (VR). 
Traditional 3D object understanding has primarily focused on pose estimation, object detection, or 3D reconstruction.
However, current approaches fail to capture the fine-grained semantics needed for reasoning about object parts, their functions, and how they can be manipulated or interacted with.
A key step toward richer understanding is to establish semantic correspondences -- estimating which points on different objects represent the same functional part.
In 2D, this problem has driven extensive research~\cite{Min19SPair,sun2021loftr,jiang2021cotr,nam2023diffmatch,mariotti2024improving}, enabling applications like image matching, retrieval, and style transfer.
Yet, 2D correspondences are inherently limited by viewpoint dependence, occlusion, and symmetry ambiguities.
We therefore propose to move beyond 2D, and towards the prediction of semantically aligned 3D locations that remain consistent across all instances of a category (as illustrated in \cref{fig:teaser}).
Unlike prior work that maps pixels into normalized canonical spaces~\cite{lin2024omninocs,wang2019normalized}, we propose to establish correspondences directly in 3D camera space, resolving fundamental ambiguities that arise in image-space matching due to occlusion, viewpoint change, and scale variation.
Formally, we define this novel task as follows:
\begin{mdframed}[backgroundcolor=eccvblue!10, linecolor=gray!60!black, roundcorner=8pt, leftmargin=0pt, rightmargin=0pt, innerleftmargin=6pt, innerrightmargin=6pt]
\textbf{\textcolor{black}{Monocular Category-level 3D correspondence:}}
Given two query and target RGB-D images $\imagesymbolquery$ and $\imagesymboltarget$ of objects from the same category, and a query 3D point $\xquery\!\in\!\mathbb{R}^3$ in the \textbf{camera space} of $\imagesymbolquery$, the task is to predict the 3D point $\xtarget\!\in\!\mathbb{R}^3$ in $\imagesymboltarget$ camera space that corresponds to the same semantic point.
\end{mdframed}
Intuitively, the task asks: \textit{if we select a semantic part on one object, where does the same part lie on another instance of the category?}
Our approach answers this question by mediating correspondence through a shared deformable template.
An overview of this camera-space correspondence setup is illustrated in \cref{fig:overview}\textcolor{red}{a}.
Unfortunately, existing benchmarks such as NOCS-Real275~\cite{wang2019normalized}, Wild6D~\cite{rodrigues2022wild6d}, OmniNOCS~\cite{lin2024omninocs}, and Omni6DPose~\cite{omni6Dpose} only provide pose annotations, segmentation, and depth, but \emph{lack category-level 3D correspondences}.
To address this gap, we introduce \textbf{\ourDataset}, a large-scale benchmark for monocular category-level 3D correspondence in camera space.
\ourDataset~covers 50 everyday object categories with 178k images and 280 unique object instances, each annotated with semantic 3D keypoints directly on CAD models that project consistently across all views.
Crucially, our annotations include \emph{amodal correspondences}---correspondences for object parts that are occluded or not visible in the image.
This capability is inspired by human reasoning~\cite{Yildirim2024-wr}, where we naturally infer the complete 3D structure of objects even under occlusion, and is essential for robotic manipulation where planning grasps and interactions requires understanding the full spatial extent of objects~\cite{xu2020learning}, not just visible surfaces.
We also explicitly support object symmetries, ensuring symmetric objects have multiple valid correspondences and avoiding unfair penalization of symmetry-equivalent predictions. 
Together, these properties address fundamental limitations of pose-focused datasets and, for the first time, enable quantitative evaluation of category-level 3D correspondence from single images.

On \ourDataset, we show that monocular category-level 3D correspondence can emerge \textbf{without explicit correspondence supervision} by constraining object instances through a shared deformable representation. 
To this end, we propose \textbf{\ours}, a framework that learns \textbf{morphable category-level shape priors} to produce semantically consistent 3D correspondences directly in camera space.
Instead of relying on a fixed representation, \ours~learns a deformable 3D template for each category that adapts to instance-specific shape variations while preserving correspondences. 
During training, our method jointly optimizes a 3D morphable prior, instance-specific shape deformations, and their 2D projection consistency.
At inference, given a single RGB-D image, \ours~predicts both the object’s 3D shape in camera space and its semantically aligned keypoints, enabling correspondence evaluation without pose normalization. 

In summary, our contributions are as follows:
\begin{enumerate}[label=(\roman*)]
    \item We identify \textbf{monocular category-level 3D correspondence in camera space} as a key next step beyond pose-centric representations toward semantically aligned 3D understanding.
    \item We introduce \textbf{\ourDataset}, the first large-scale benchmark for category-level 3D correspondence, comprising 178k images across 50 household categories and 280 instances, with mesh-based keypoint annotations, amodal correspondences, and explicit symmetry labels.
    \item We propose \textbf{\ours}, a framework that learns \textbf{morphable category-level shape priors} to establish semantically consistent 3D correspondences directly in camera space.
    \item We demonstrate that \ours~substantially outperforms existing baselines on \ourDataset, establishing a new paradigm for \textbf{correspondence-level 3D object understanding}.
\end{enumerate}

\section{Related work}

\emph{2D Semantic Correspondence.}
2D correspondence has advanced from local descriptors and dense flows 
(\eg, SIFT~\cite{Lowe04SIFT}, DAISY~\cite{Tola10DAISY}, SIFT Flow~\cite{Liu11SIFTFlow}, DeepFlow~\cite{Weinzaepfel13DeepFlow}) to transformer-based self-supervised features~\cite{caron2021emerging,zhou2021ibot,oquab2023dinov2,zhang2023tale}, which exhibit emergent semantic alignment and achieve strong results on benchmarks like SPair-71K, PF-PASCAL, and TSS~\cite{Min19SPair,Ham16,li2023simsc}.  
Dedicated matchers such as LoFTR, COTR, DiffMatch~\cite{sun2021loftr,jiang2021cotr,nam2023diffmatch}, and spherical-map approaches~\cite{mariotti2024improving,duenkel2025diysc} further improve dense matching.
While highly effective, these approaches remain limited to the image domain and do not predict 3D canonical coordinates or enforce semantic consistency across instances in 3D space.

\emph{3D Keypoint and Correspondence Methods.}
Prior work explored correspondence mapping in the 3D domain through keypoint detection and surface mapping.
KeypointNet~\cite{KeypointNet2020} introduced a large-scale dataset for learning category-consistent 3D keypoints, while others~\cite{keypointdeformer2021,neuralcage2020} leverage keypoints for cage-based deformations and shape control.
Canonical surface mapping~\cite{canSurfMap2019abhinav} establishes correspondences by predicting UV coordinates on canonical templates, and Mesh R-CNN~\cite{meshrcnn2019} jointly predicts mesh reconstructions with instance segmentation from 2D images.
Recent semantic alignment methods~\cite{cewu22understandingsemantic,cewu2020humancorr,semalign3d2025} explore learning consistent correspondences across categories and human poses in 3D.
DenseMatcher~\cite{zhu2024densematcher} extends matching to the mesh domain via functional maps, projecting multiview features onto 3D geometry.
However, these approaches have fundamental limitations: 
KeypointNet~\cite{KeypointNet2020}, Keypointdeformer~\cite{keypointdeformer2021}, \cite{neuralcage2020}, and DenseMatcher~\cite{zhu2024densematcher} require ground-truth 3D meshes as input;
methods like~\cite{cewu22understandingsemantic,cewu2020humancorr,keypointdeformer2021} operate exclusively in 3D space without bridging to image-based features;
and critically, none provide large-scale evaluation benchmarks with explicit handling of occlusion and symmetry.
These limitations prevent their applicability to real-world scenarios where RGB(-D) images are predominantly available.

\emph{Morphable Models and Shape Priors.}
Morphable models achieve category-level understanding by capturing intra-class shape variability through deformable canonical templates.
Classic work focused on faces and human bodies (\eg, 3D Morphable Models~\cite{blanz1999morphable}, SMPL~\cite{loper2015smpl}), establishing the foundation for template-based shape modeling.
Recent approaches~\cite{Neverova20,SHIC,Common3D,MeshUp} extend these ideas to more diverse object classes using learned deformations or diffusion-guided generation.
Deformation-based methods~\cite{groueix2018b,wang2018pixel2mesh,hee2020shapepriordeform} map instances to template meshes using neural networks, while template-free approaches~\cite{novotny2019c3dpo} learn canonical coordinate systems without relying on a single exemplar.
More recent work leverages foundation models for semantic alignment across categories~\cite{Neverova20,SHIC}, where semantically corresponding parts map to consistent representations.
Domain-specific efforts have also addressed human bodies~\cite{Guler18} and a range of animals~\cite{xu2023animal3d}.
Despite this progress, generalizing morphable models to diverse everyday objects with consistent 3D correspondences across instances remains an open challenge, especially for methods that operate only from image inputs.

\emph{Benchmarks for Category-Level 3D Understanding.}
To the best of our knowledge, there exists no dataset that enables category-level 3D correspondence evaluation from monocular images.
Prior works~\cite{wu2023magicpony} lift 2D images from domain-specific datasets~\cite{CUB_dataset2022,wu2023dove} to 3D using multi-view consistency but lack 3D evaluation benchmarks.
Large-scale 3D shape collections such as ShapeNet~\cite{Chang15} and ModelNet~\cite{Wu15} provide CAD meshes, while ShapeNetPart~\cite{Yi16} and PartNet~\cite{Mo19} add part-level labels, but these lack consistent point-level correspondences across instances.
Pose-focused datasets like Omni6DPose~\cite{omni6Dpose}, CO3D~\cite{co3d}, Pix3D~\cite{pix3d}, Pascal3D+~\cite{xiang2014beyond}, and Omni3D~\cite{brazil2023omni3d} provide pose annotations in realistic scenes but do not supply semantic, amodal, or point-level correspondences across diverse instances.
NOCS datasets~\cite{wang2019normalized,lin2024omninocs} introduced normalized coordinate spaces for pose estimation but are not designed for evaluating category-level correspondences, as described in \cref{suppseclimitation_existing_bench}.
DenseCorr3D~\cite{zhu2024densematcher} takes a valuable step with part-level mesh annotations and functional-map evaluation, but operates exclusively in 3D with pre-reconstructed meshes. Thus, current 3D benchmarks do not bridge the gap between 2D-based and 3D correspondence methods.\\
In contrast, \ourDataset~is explicitly designed for category-level 3D correspondence evaluation from monocular images, featuring 3D keypoints shared across all instances within 50 object categories, with amodal labels for occluded regions and explicit symmetry handling.
This addresses a fundamental gap in current datasets and enables quantitative evaluation of correspondence-based 3D object understanding in camera space.

\section{The \ourDataset Benchmark}\label{sec:dataset}
\begin{figure*}[t]
    \centering
    \includegraphics[width=\linewidth]{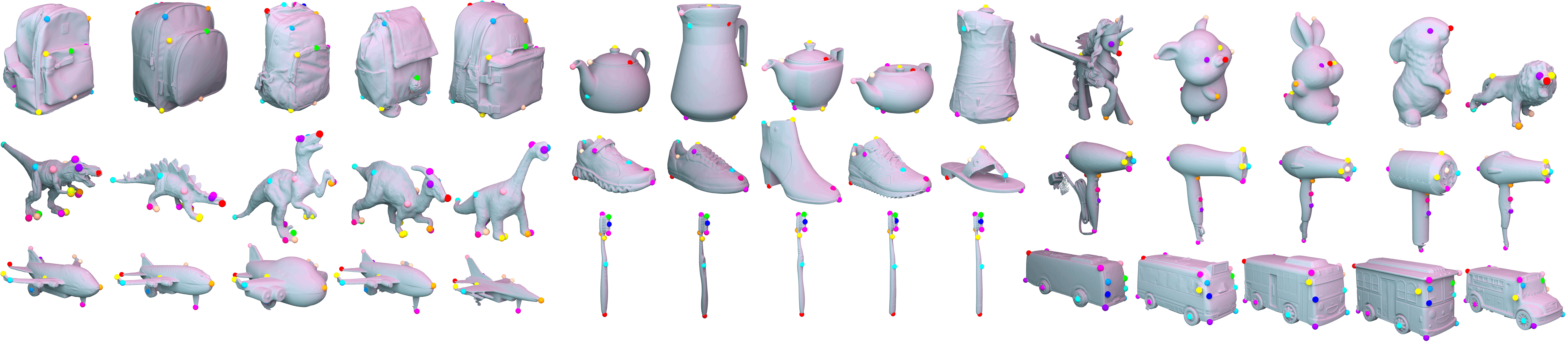}
    \caption{\textbf{Dataset Overview.} We annotate up to 19 3D keypoints directly on CAD meshes for 5--13 instances per category, covering 50 common household object classes. 
    The keypoints are chosen to be semantically consistent and shared across all instances within each category
    We visualize a subset of these annotations across several categories to highlight their cross-instance and cross-shape consistency.
    Visualizations for the full dataset are provided in \cref{suppsec:annotation}.}
    \label{fig:dataset_overview}
    \vspace{-1em}
\end{figure*}

\emph{Motivation.} We introduce the first benchmark for category-level correspondences in 3D camera space, unlike prior datasets that focus exclusively on correspondences in either 2D camera space~\cite{Min19SPair,Ham16, sun2023misc210k, CUB_dataset2022, wu2023dove} or 3D object space~\cite{zhu2024densematcher}. On the one hand, compared to reasoning in 3D object space, advancing monocular methods at estimating in 3D camera space, \textbf{removes the need for ambiguous object-centric spaces}, whereby neither the center nor the scale is well-defined. Moreover, compared to estimation in 2D camera space the \textbf{3D camera space has several critical advantages}: a) the evaluation of amodal correspondences, b) modeling object symmetries explicitly, and c) enforcing methods to perform 3D over 2D reasoning. 
Importantly, \ourDataset~is designed as a \emph{test-only benchmark}: keypoints annotations are used exclusively for evaluation.

\begin{table}[t]
\caption{
\textbf{Comparison to existing correspondence datasets.}
Prior benchmarks evaluate in either 2D camera or 3D object space. \textbf{\ourDataset} is the first to target 3D camera space, enabling amodal evaluation across 50 classes.
}
\label{tab:dataset_stats2d}
\centering
\vspace{-1em}
\footnotesize
\setlength{\tabcolsep}{5pt}
\begin{tabular}{lrrrlcc}
Dataset        & pairs  & classes  & input & eval. space & symmetry & occlusion \\
\midrule
Pascal-Parts~\cite{Chen14DetectWhatYouCan}& 4k  & 20  & 2D & 2D camera & \xmark & \xmark\\
PF-Pascal~\cite{Ham16}                   & 2k  & 20  & 2D & 2D camera & \xmark & \xmark\\
Spair71k~\cite{Min19SPair}               & 71k & 18  & 2D & 2D camera & \xmark & \cmark\\
KeypointNet~\cite{KeypointNet2020}       & N/A & 16  & 3D & 3D object & \xmark & \xmark\\
CPNet~\cite{cewu2020humancorr}           & N/A & 25  & 3D & 3D object & \cmark & \xmark\\
DenseCorr3D~\cite{zhu2024densematcher}   & N/A & 23  & 3D & 3D object & \cmark & \xmark\\
\textbf{\ourDataset}                     & \textbf{178k} & \textbf{50} & \textbf{2.5D} & \textbf{3D camera} & \cmark & \cmark\\
\end{tabular}
\vspace{-2em}
\end{table}

\emph{Task definition.} 
Given two RGB-D images $\imagesymbolquery$ and $\imagesymboltarget$ depicting objects from the same category, and a query 3D point $\xquery\!\in\!\mathbb{R}^3$ in the camera space of $\imagesymbolquery$, the task is to predict the corresponding 3D point $\xtarget\!\in\!\mathbb{R}^3$ in the camera space of $\imagesymboltarget$ that represents the same semantic part of the object.
Formally, it can be expressed as a mapping 
$f:(\xquery,\imagesymbolquery, \imagesymboltarget)\rightarrow\xtarget$. 
The evaluation is performed using the Euclidean distance between the groundtruth target point $\xtarget$ and the predicted target point $\xtargetproj$, defined as 
$d(\xtargetproj, \xtarget)=\left\lVert\xtargetproj - \xtarget\right\rVert_2$.
The performance of a model is measured by computing the percentage of correctly predicted points within a given threshold on the euclidean distance (\eg, PCK@0.1), using the largest of, width $w$, height $h$, and depth $d$ of the object's 3D bounding box, as: $d(\xtargetproj, \xtarget) < 0.1 \cdot \mathrm{max}(h, w, d)$.
This follows the conventions of other monocular 2D correspondence benchmarks~\cite{Min19SPair,Ham16, sun2023misc210k, CUB_dataset2022, wu2023dove}, where the maximum width and height of the 2D bounding box are used to normalize the distance and compute PCK.
Further discussion of correspondence evaluation, including the distinction between modal and amodal settings, is provided in \cref{suppsec:correspondence_evaluation}. 

\emph{\ourDataset.} 
We build our dataset on Omni6DPose~\cite{omni6Dpose}, a large-scale synthetic dataset designed for category-level pose estimation in crowded scenes. We crop the images to obtain 178k test and 2.6M train images across 50 categories. We find 178k image pairs, by choosing a random image for each test image, which contains another instance. 
We specifically leverage Omni6DPose synthetic subset, which provides photo-realistic renderings with high-quality CAD models of real object instances, natural lighting, cluttered scenes, and realistic occlusions.
Unlike the real subset which contains limited instance diversity (typically 1--2 instances per category) and repetitive scene layouts due to video-frame extraction, the synthetic data provides greater scale and instance diversity, which is beneficial for learning robust category-level correspondences.
We select 50 everyday object categories spanning household items (mugs, bottles, remotes), food items (fruits, vegetables), toys (cars, planes, animals), and accessories (backpacks, shoes, wallets), chosen to maximize shape diversity and practical relevance for robotic manipulation.
For each category, between 2 and 19 semantic 3D keypoints are annotated directly on CAD meshes (see \cref{fig:dataset_overview}).

\emph{Keypoint Annotation Protocol.} 
Keypoints must be shared across all instances of a category and are selected to be geometrically distinctive and semantically meaningful~\cite{SuwajanakornSTN18}---marking corners, edges, handle centers, or other salient structural features rather than arbitrary surface points.
This ensures that annotations are both reliably localizable and transferable across instances.
To ensure annotation quality and consistency, we employ a rigorous protocol (more details in \cref{suppsec:mesh_annot}) involving two annotators\footnote{Annotators are trained on best practices for selecting geometrically distinct and semantically meaningful keypoints that are localizable and consistent across instances.} independently annotate the same set of meshes using an interactive 3D tool. Following this process, a two-stage merging process is applied including an initial \textit{automatic merging} step which computes mutual nearest-neighbor matches between the two annotation sets across all instances based on distance (5\%-threshold of object bounding-box diagonal) and consistency (pairs of keypoints are matched consistently), annotations are considered accepted or undecided. Then a second \textit{manual merging} step is performed for undecided keypoints. Annotators use an interactive 3D viewer displaying multiple instances side-by-side to manually resolve ambiguities: accepting, rejecting, splitting, or merging annotations based on semantic and geometric consistency.
The entire annotation process took approximately 65h across both annotators, yielding a total set of 2329 3D keypoint annotations on meshes by annotating between 2 and 19 keypoints per instance.
Once keypoints are annotated on 3D meshes, we leverage ground-truth poses from Omni6DPose~\cite{omni6Dpose} to automatically project them into all rendered views, generating consistent 2D--3D correspondences across 178k pairs of images with minimal additional manual effort.
This mesh-centric strategy offers three key advantages: (i) it enforces \emph{semantic consistency} across all views and instances, (ii) it naturally provides \emph{amodal} labels for occluded regions, and (iii) it efficiently scales a compact set of 3D annotations into a large-scale benchmark spanning 178k pairs across 50 categories and 280 instances.
The resulting benchmark inherits the visual realism of Omni6DPose, featuring natural lighting, cluttered scenes, and partial occlusions.

\emph{Symmetry.} 
Many everyday objects exhibit geometric symmetries that introduce fundamental ambiguities in correspondence.
For instance, a cylindrical mug body is rotationally symmetric---any point on the rim can rotate to any other without changing the object's shape.
To the best of our knowledge, existing semantic correspondence benchmarks have not addressed symmetries, as they operate purely in 2D where such geometric constraints are difficult to define.
By leveraging 3D annotations, \ourDataset~explicitly handles \emph{discrete} and \emph{continuous} symmetries, ensuring that geometrically equivalent predictions are not unfairly penalized.
Symmetry is handled by treating all points on the orbit generated by rotations around the symmetry axis as valid correspondences.
This yields a fair metric that respects the inherent geometric ambiguities in real-world objects and enables robust evaluation of category-level correspondence methods.
More details are provided in \cref{suppsec:correspondence_evaluation}.

\section{Method}\label{sec:method}
Our goal is to recover category-level 3D correspondences directly in camera space from monocular RGB-D observations. 
To achieve this, we introduce \textbf{Morpheus}, a model that, from a single image, predicts a 6D object pose and a deformable 3D shape whose semantic structure remains consistent across object instances. 
The central idea of Morpheus is to represent all objects within a category as \emph{identity-preserving deformations} of a shared template mesh. 
Because template vertices maintain persistent identities during deformation, semantic correspondences arise naturally: points associated with the same template vertex correspond to the same semantic part across instances.
We start by describing how to predict 3D correspondences in camera space in \cref{sec:inference}. Subsequently, we explain our architecture in \cref{sec:morphable_priors}, and finally we elaborate on the objectives in \cref{sec:method_obj}.
 \begin{figure*}[t]
    \centering
    \includegraphics[width=\textwidth]{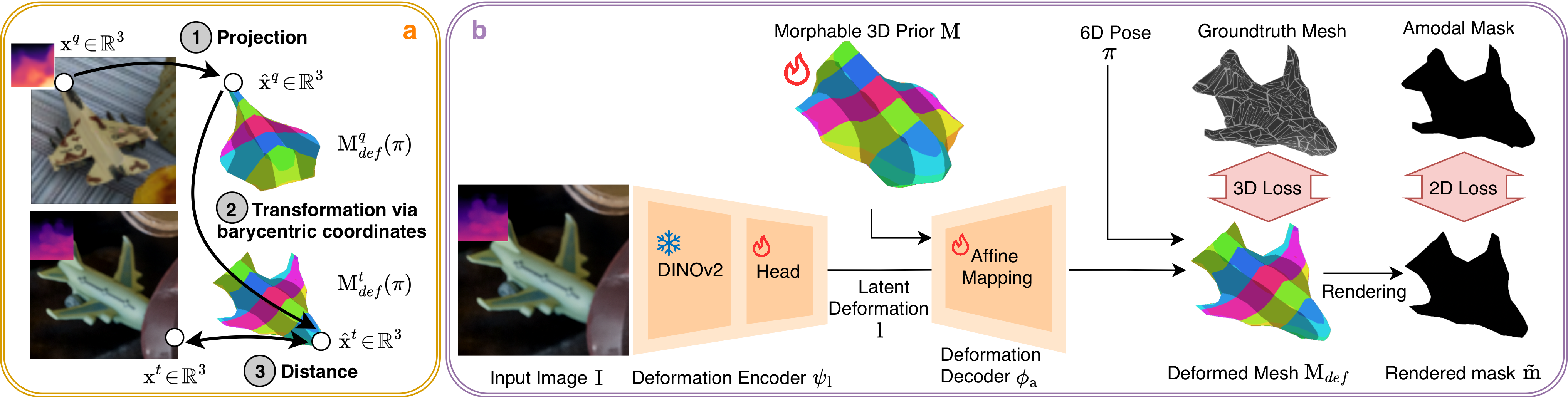}
    \caption{
    \textbf{\textcolor{pipelineorange}{(a)} Monocular category-level 3D correspondence.} Given a query point $\xquery \in \mathbb{R}^3$, we project it onto the deformed query mesh $\meshdefquery$ and encode its location as barycentric coordinates. Since query and target instances share the same mesh topology, these coordinates transfer directly to $\meshdeftarget$, yielding the corresponding point $\xtarget \in \mathbb{R}^3$.
\textbf{\textcolor{pipelinepurple}{(b)} Pipeline.} Given an RGB-D image, the deformation encoder $\featencl$ predicts a latent code $\instlatent$ that drives the decoder $\affinef$ to adapt the category shape prior to the observed instance. The deformed mesh is placed in camera space using the predicted 6D pose. Training uses amodal 2D and 3D losses together with pose supervision.
}
    \label{fig:overview}
    \vspace{-1em}
\end{figure*}

\emph{Notation}
We denote a mesh as $\mesh\!=\!\{\verts, \edges\}$, with vertices $\verts\!=\!\{\vei\!\in\!\mathbb{R}^3\}_{i=1}^{|\verts|}$ and edges $\edges = \{ (\vei, \vej)_{e} \}_{e=1}^{|\edges|}$.
For correspondence tasks, $-^\query$ and $-^\target$ distinguish query and target elements (\eg, $\meshquery$ and $\meshtarget$).
We denote a deformed mesh as $\meshdef$, and its transformation into camera space with pose $\pose$ as $\meshdef(\pose)$.
\subsection{Mesh-based 3D Correspondence Prediction} \label{sec:inference}
\ours establishes correspondences by mediating all predictions through a shared deformable template.
For each RGB-D image $I$, the model predicts:
(i) an instance-specific deformation of the template mesh, and
(ii) a 6D pose $\omega$ estimated from pretrained pose diffusion~\cite{omni6Dpose}.
The deformed mesh is then transformed into camera space as $M_{\mathrm{def}}(\omega)$.
Given a query–target image pair $(I^q, I^t)$, we obtain their posed meshes as
$M^q_{\mathrm{def}}(\omega^q)$ and
$M^t_{\mathrm{def}}(\omega^t)$.

\emph{3D Correspondence Prediction via Mesh Transfer.}
A query 3D point $\mathbf{x}_q$ is first projected onto the surface of the query mesh, producing a surface point $\hat{\mathbf{x}}_q$. 
We represent this point using barycentric coordinates with respect to the underlying mesh face.
Because both instances share identical mesh topology, these barycentric coordinates define a category-level surface identifier.
The same identifier is then transferred onto the target mesh, yielding the predicted correspondence $\hat{\mathbf{x}}_t$ in the target camera space (Fig.~3a).
Thus, monocular 3D correspondence estimation reduces to predicting the pose and deformations of a shared template mesh rather than directly matching points between images.

\subsection{3D Morphable Priors}  \label{sec:morphable_priors}
A central component of \ours~is the \emph{3D morphable prior}, which models all instances of a category as deformations of a shared canonical template, hence enabling semantically consistent correspondences across instances.
It consists of a canonical mesh capturing the common topological structure of a category, along with a learned deformation model that adapts it to individual instances. 
We refer to the model as a \emph{prior} because all predictions are constrained to be deformations of this canonical representation. 
Since each vertex of the template retains its identity across deformations, semantic correspondences are preserved by design: observations mapped to the same template vertex correspond to the same semantic part across instances.
This converts 3D correspondence estimation into a pose and deformation estimation problem.
Unlike prior morphable-model approaches designed for reconstruction, our formulation leverages persistent template vertex identity as the fundamental mechanism enabling monocular category-level 3D correspondence.

\emph{Canonical Shape Representation.} Traditional mesh-only representations are often fragile and difficult to optimize directly, typically requiring manual interventions such as remeshing \cite{goel2022differentiable,yang2021lasr}. 
To overcome this limitation, we employ a hybrid volumetric mesh representation~\cite{shen2021deep}. This integrates the strengths of implicit and explicit 3D models.  
Concretely, the category-level shape is represented as a signed distance field $\sdff$, providing flexibility to model intricate geometries.  
Through Differentiable Marching Tetrahedra~\cite{shen2021deep}, the SDF is efficiently transformed into a mesh in a differentiable manner by evaluating SDF values on a tetrahedral grid. 
This formulation enables the use of mesh-based priors and regularizations, such as enforcing rigidity constraints during deformation learning. 

\emph{Instance-Specific Deformations.}  
To adapt this canonical mesh to specific instances, we learn an affine deformation field, following \cite{zheng2021deep}.  
Unlike \cite{zheng2021deep}, where deformations are applied directly to the signed distance field, we act on the template mesh vertices \cite{wu2023magicpony}. This avoids repeatedly extracting meshes for each instance and is thus more computationally efficient.  
Formally, we define an affine mapping $\affinef: \mathbb{R}^3 \times \instlatentspace \to \mathbb{R}^3$, which displaces each vertex $\vv$ individually according to the instance-specific latent code $\instlatent$:  
\begin{equation}
    \affinef(\vv, \instlatent) = \affinefscale(\vv, \instlatent) \odot \vv + \affineftransl(\vv, \instlatent),
\end{equation}  
where $\affinefscale,\, \affineftransl : \mathbb{R}^3 \times \instlatentspace \rightarrow \mathbb{R}^3$ are produced by an MLP that takes both the vertex coordinate $\vv$ and the latent code $\instlatent$ as input.  
The latent code $\instlatent = \featencl(\imagesymbol)$ itself is computed from the input image $\imagesymbol$ by a deformation encoder $\featencl$ built from a DINOv2 backbone with a light convolutional head.
This code parametrizes vertex-wise displacements, enabling the mesh to morph into the observed instance while preserving semantic alignment. 
The resulting instance mesh is $\meshdef(\imagesymbol) = \{\vertsdef(\imagesymbol), \edges \}$, where each deformed vertex is given by $\vertsdef(\imagesymbol) = \{ \affinef(\vei, \featencl(\imagesymbol)) \}_{i=1}^{|\verts|}$.  
For simplicity, we simply rewrite it as $\meshdef = \affinef(\mesh, l)$.
Through the deformation, vertices maintain consistent identities, enabling category-level correspondence prediction without explicit correspondence supervision.

\subsection{Training Objectives}\label{sec:method_obj}
\ours{} is trained using geometric supervision that encourages all object instances to explain observations through a shared morphable category-level prior. 
Importantly, no explicit correspondence supervision is used during training. 
Instead, semantic alignment emerges implicitly because every instance must deform the same canonical template while remaining consistent with observed data.
We jointly optimize the encoder, decoder and morphable prior using reconstruction and regularization objectives. 
Specifically, training enforces consistency between the deformed mesh and the input observations (\cref{fig:overview}\textcolor{red}{b}) through (i) 2D mask-based reconstruction, (ii) 3D geometry alignment, and (iii) deformation regularization. 
Together, these objectives encourage the model to learn category-consistent canonical structure while preserving instance-specific shape variation.
In contrast to prior work~\cite{wu2023magicpony}, we additionally provide 6D pose supervision to stabilize optimization and reduce local minima. 
Furthermore, amodal 2D supervision and 3D geometric losses improve robustness under occlusion by encouraging reasoning about both visible and occluded object regions.

\emph{2D Loss.}  
We first supervise using amodal object masks. Given the predicted mask $\maskpred(\meshdef, \imagesymbol, \pose)$ rendered from the deformed mesh $\meshdef$ under pose $\pose$, we compare against the ground-truth (GT) amodal mask $\mask$ with a pixel-wise MSE:
\begin{equation}
    \lossmask(\meshdef, \imagesymbol, \pose, \mask) 
    = \big\lVert \maskpred(\meshdef, \imagesymbol, \pose) - \mask \big\rVert^2.
\end{equation}
Additionally, we encourage overlap with the distance transform of the ground-truth amodal mask $\maskdt$:
\begin{equation}
    \lossmaskdt(\meshdef, \imagesymbol, \pose, \mask) 
    = - \maskpred(\meshdef, \imagesymbol, \pose) \odot \maskdt,
\end{equation}
with $\maskdt$ encoding the distance of each pixel inside the mask to the silhouette boundary, while pixels outside the mask are zero, which prevents disconnected parts from emerging when fitted across diverse instances.

\emph{3D Loss.}  
For accurate 3D instance reconstruction, we use a Chamfer distance between the deformed mesh vertices $\vertsdef$ and the GT mesh vertices $\vertsdefgt$.
\begin{equation}
\lossrec(\imagesymbol, \meshdef, \meshgt)
= \tfrac{1}{|\vertsdef| + |\vertsdefgt|} \Big(
\!\sum_{\vv_i \in \vertsdef} \!\lVert \vv_i - \vv'_{\vnneuc(\vv_i)}\rVert\!
+ \!\sum_{\vv'_i \in \vertsdefgt} \!\lVert \vv'_i - \vv_{\vnneuc(\vv'_i)}\rVert\! \Big),
\end{equation}
where $\vnneuc$ denotes the nearest neighbor operator.  

\emph{Template and Deformation Regularization.}
Following~\cite{gropp2020implicit}, we enforce the SDF property with the Eikonal loss $\losseik$, penalize large deformation with an $\ell_2$ term: $\lossregdef$, and encourage smoothness with an edge-based regularization $\lossregdefsm$ \cite{zheng2021deep}. Their definitions are provided in \cref{suppsec:losses}.

\emph{Training Loss.}  
Our optimization proceeds in two stages.  
First, we refine the category-level template using only geometric terms:
\begin{equation}
\label{eq:loss_1}
    \mathcal{L}_\text{geo} = 
    \glossrec \lossrec + 
    \glossmask \lossmask + 
    \glossmaskdt \lossmaskdt + 
    \glosseik \losseik.
\end{equation}
After convergence, we learn the instance deformations with the extended loss:
\begin{equation}
\label{eq:loss_2}
    \mathcal{L}_\text{geo-reg} = 
    \mathcal{L}_\text{geo} + 
    \glossregdef \lossregdef + 
    \glossregdefsm \lossregdefsm.
\end{equation}

\begin{figure*}[t]
    \centering
    \includegraphics[width=\linewidth]{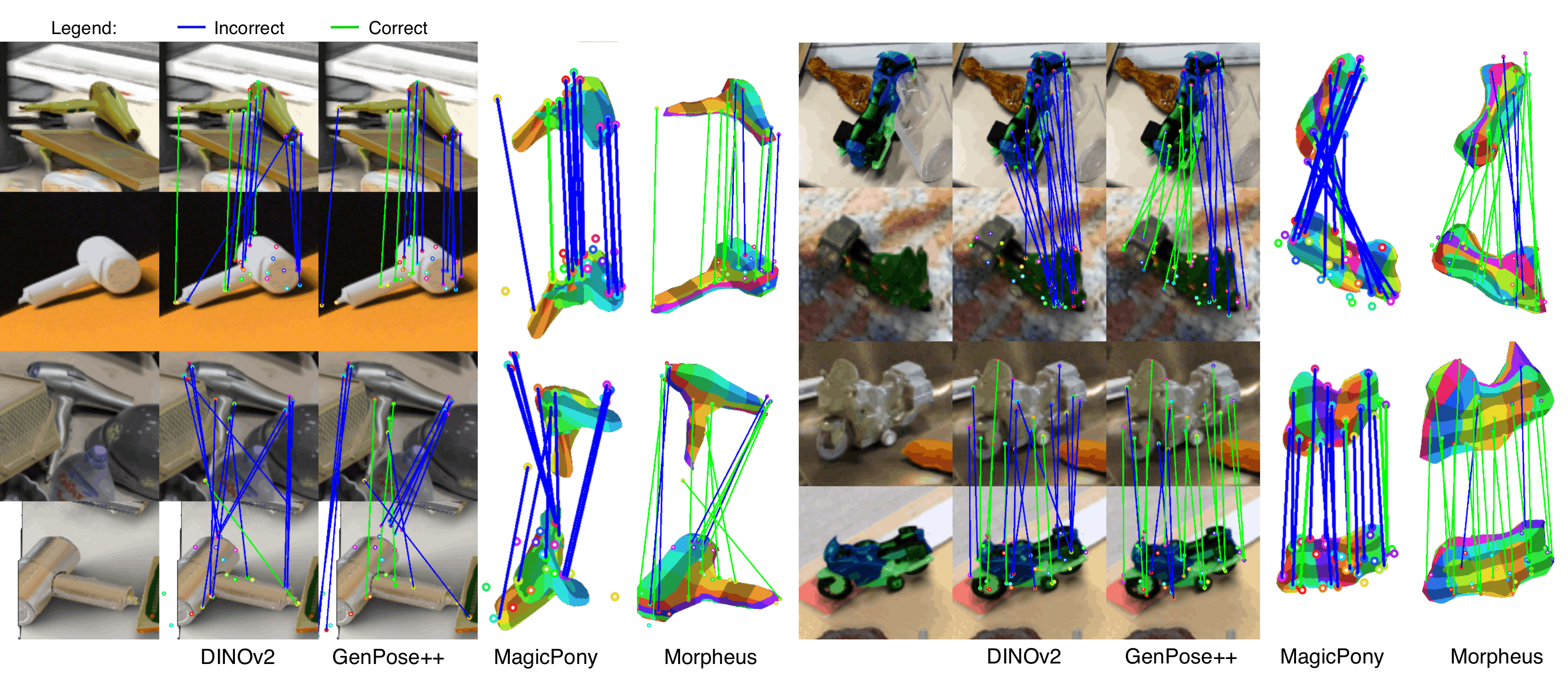}
    \caption{\textbf{Qualitative results.} We compare 2D feature matching method DINOv2~\cite{oquab2023dinov2}, with 3D space matching methods GenPose++~\cite{omni6Dpose}, MagicPony~\cite{wu2023magicpony}, and \ours. For DINOv2 and GenPose++ we visualize the 2D correspondences. For MagicPony and \ours, we visualize the predicted deformed meshes in camera space, along with overlaid correspondence lines (see \cref{suppsec:visualization}). 
    MagicPony's predictions may appear visually plausible when projected in 2D but are often incorrect in 3D (\eg, bottom-right), as 2D supervision alone does not constrain the 3D structure to be consistent. Additionally, pose-aware training results in higher consistency for semantic parts across different viewpoints. Note that DINOv2 often confuses parts, and GenPose++ may predict points outside the object due to its rigid shape assumption.}
    \label{fig:qualitative}
\vspace{-1em}
\end{figure*}
\section{Experiments}\label{sec:experiments}
\begin{table}[t]
\centering
\footnotesize
\caption{\textbf{PCK@0.1 results} for 2D, 3D modal, and 3D amodal correspondences on a subset of \ourDataset. \ours~outperforms all 2D correspondence methods (DINOv2\cite{oquab2023dinov2}, $\text{MagicPony}_{\text{2D}}$\cite{wu2023magicpony}, NOCS\cite{wang2019normalized} and 3D methods (GenPose++ (GP++)\cite{omni6Dpose}, MagicPony\cite{wu2023magicpony}, and \ours).
$^\star$2D predictions lifted to 3D via depth; amodal evaluation is not applicable (occluded points have no depth).
}
\resizebox{\columnwidth}{!}{%
\setlength{\tabcolsep}{5pt}
\begin{tabular}{l|ccccccc|ccccccc}
\toprule
Method & \Backpack & \faPlane & \faShip & \faBus & \faBreadSlice &\faMotorcycle & mean & \Backpack & \faPlane & \faShip & \faBus & \faBreadSlice &\faMotorcycle & mean\\
\hline 
\multicolumn{1}{c|}{\cellcolor{gray!20}}&\multicolumn{7}{c|}{\cellcolor{gray!20}\textbf{2D}} & \multicolumn{7}{c}{\cellcolor{gray!20}\textbf{3D Modal}} \\
DINOv2$^\star$        & 7.0  & 15.2 & 17.1 & 13.3 & 14.0 & 10.6 & 22.9 & 5.7  & 9.2  & 7.5  & 14.4 & 16.4 & 11.0 & 24.4 \\
$\text{MagicPony}_{\text{2D}}$$^\star$ & 6.4  & 7.7  & 8.8  & 7.2  & 22.9 & 9.1  & 15.7 & 3.9  & 3.1  & 2.7  & 4.7  & 22.7 & 10.1 & 14.0 \\
NOCS$^\star$         & 27.2 & 20.7 & 14.0 & 42.6 & 23.7 & 16.6 & 26.7 & 6.5  & 13.5 & 4.5  & 34.6 & 24.0 & 7.4  & 26.4 \\
\hline
GP++          & 37.0 & 28.8 & 20.5 & 50.2 & 30.0 & 26.7 & 36.3 & 22.9 & 14.9 & 12.9 & 38.5 & 27.9 & 27.5 & 37.0 \\
MagicPony+GP++    & 4.8  & 7.2  & 4.1  & 4.2  & 22.1 & 8.1  & 10.7 & 2.5  & 2.1  & 1.1  & 0.3  & 14.7 & 4.1  & 7.5  \\
\ours w/o Def. & 39.9 & 32.0 & 22.5 & 51.8 & 34.2 & 29.5 & 39.1 & 25.2 & 16.8 & 17.2 & 44.5 & 35.2 & 27.8 & 40.2 \\
\ours    (Ours)       & \textbf{40.9} & \textbf{34.8} & \textbf{28.1} & \textbf{57.1} & \textbf{36.5} & \textbf{31.3} & \textbf{41.2} & \textbf{26.0} & \textbf{23.6} & \textbf{19.9} & \textbf{49.2} & \textbf{38.8} & \textbf{33.8} & \textbf{43.7} \\
\doublehline
\multicolumn{1}{c|}{\cellcolor{gray!20}}&\multicolumn{7}{c|}{\cellcolor{gray!20}\textbf{3D Amodal}} & \multicolumn{7}{c}{\cellcolor{gray!20}\textbf{3D (Modal + Amodal)}} \\
GP++         & 17.1 & 19.2 & 14.8 & 36.7 & 27.3 & 15.0 & 32.9  & 18.8                 & 18.2                 & 14.4                 & 37.1                 & 27.5                 & 17.9                 & 34.3                 \\
MagicPony+GP++    & 0.7 & 2.1 & 0.9 & 1.1  & 9.1  & 1.6  & 7.1 & 1.2                  & 2.1                  & 0.9                  & 0.9                  & 10.8                 & 2.2                  & 7.1                  \\
\ours w/o Def. & 21.6 & 23.1 & 16.3 & 40.3 & 34.9 & 17.3 & 37.8 & 22.7                 & 21.6                 & 16.5                 & 41.3                 & 35.0                 & 19.8                 & 38.4 \\
\ours  (Ours)         & \textbf{22.8} & \textbf{26.7} & \textbf{21.3} & \textbf{47.5} & \textbf{39.4} & \textbf{19.0} & \textbf{40.8} & \textbf{23.7}                 & \textbf{26.0}                 & \textbf{21.0}                 & \textbf{47.9}                 & \textbf{39.2}                 & \textbf{22.5}                 & \textbf{41.5}                 \\
\bottomrule
\end{tabular}%
}
\label{tab:all_corres}
\vspace{-1em}
\end{table}

We evaluate \ours on the proposed \ourDataset benchmark, focusing on its ability to recover \emph{category-level 3D correspondences}. 
We compare \ours with strong 2D correspondence baselines such as NOCS~\cite{wang2019normalized} and DINOv2~\cite{oquab2023dinov2}, as well as 3D space matching methods such as MagicPony~\cite{wu2023magicpony} and GenPose++~\cite{omni6Dpose}. We first provide experimental details in \cref{sec:exp_details}. We describe all baselines in \cref{sec:exp_baselines} and finally compare with prior work in \cref{sec:exp_compare}.     

\subsection{Experimental Details} \label{sec:exp_details}
\ours uses a pretrained ViT-S DINOv2 image encoder~\cite{oquab2023dinov2} as backbone, and a pretrained 6D pose diffusion network~\cite{omni6Dpose}.
From an input resolution of $448^2$, the backbone maps to a $32^2$ feature map. The deformation encoder is implemented as a ResNet head~\cite{resnet} that aggregates multi-scale feature maps with bottleneck blocks to produce refined latent deformation $\mathrm{l}$.
The deformation decoder is a coordinate-conditioned MLP that fuses 3D point embeddings with latent deformation to predict deformations.
To learn the initial template shape, we train each category-specific morphable model using the Adam optimizer \cite{kingma2015adam} with a learning rate of $10^{-4}$ and a batch size of $30$. 
Training proceeds in two stages: (i) 20 epochs optimizing the loss in \cref{eq:loss_1}, and (ii) 10 further epochs optimizing the extended loss in \cref{eq:loss_2}, which includes deformation regularizers. Training on a NVIDIA RTX 2080 takes about 12h. 

\emph{2D and 3D Metrics.}
For our benchmark, we use the percentage of correct keypoints (\ie, PCK@$0.1$) as described in \cref{sec:dataset}. We differentiate between 2D evaluation, where the distance is measured in pixel space and the threshold depends on the 2D bounding box, and 3D evaluation, where the distance is measured in camera space and the threshold depends on the 3D bounding box. In 3D, we further distinguish between modal correspondences (where the keypoint is visible in both images) and amodal correspondences (where one keypoint is occluded). 
Ambiguities due to object symmetries can lead to multiple valid correspondences, which we handle separately.
We provide more details in \cref{suppsec:correspondence_evaluation}.

\subsection{Baselines}\label{sec:exp_baselines}
Given our newly defined task, we made every effort to identify competitive baselines capable of processing RGB-D input data and producing predictions in both 2D and 3D domains.
We first compare against 2D feature-matching baselines such as NOCS~\cite{wang2019normalized} and DINOv2~\cite{oquab2023dinov2}, where each pixel is represented by a feature vector in $\mathbb{R}^d$ and matched to its nearest neighbor in the target image.
In this context, predicted NOCS coordinates are treated as features in $\mathbb{R}^3$.
For MagicPony, we render its canonical-space coordinates and use the rendered results as a 2D feature-matching baseline (denoted as $\text{MagicPony}_{\text{2D}}$).
Using the target image’s depth map, the predicted 2D pixels can be reprojected into 3D, enabling 3D \emph{modal} correspondences.
However, since occluded regions are not visible, the 3D \emph{amodal} correspondence task cannot be solved using any 2D baseline.
Additionally, we compare with 3D space matching baselines such as GenPose++ and MagicPony. MagicPony also uses a 3D morphable prior; thus, the template mesh can be used to match points in 3D as explained in \cref{sec:inference}. In contrast, GenPose++ does not predict a 3D shape, but only a 6D pose. However, we can transform the query points from camera space into the normalized object-centric space using the inverse query camera pose, and further to the target camera space using the target camera pose. 
As MagicPony learns canonical space and camera space entangled together, an external orientation estimator cannot be applied as its canonical space would not match the learned canonical orientation. However, we still require the translation and object scale from GenPose++ to obtain 3D correspondence predictions.

\subsection{Comparison with Prior Work}\label{sec:exp_compare}

Overall, \cref{tab:all_corres} shows that \ours~sets a new state of the art on both 2D and 3D correspondence metrics. 
\cref{fig:qualitative} illustrates qualitative predictions of \ours. 

\emph{Occlusions.} 
2D feature matching methods (\eg, NOCS, DINOv2) cannot handle occlusions by design, and cannot evaluate them in the 3D amodal setting. In \cref{fig:qualitative}, we also observe how DINOv2 matches the back of the hairdryer with the front of another one, which is truncated in the target image. 
Furthermore, we observe qualitatively that MagicPony fails to correctly reconstruct occluded parts, as seen for the hairdryer in \cref{fig:qualitative}. In contrast, \ours~successfully reconstructs occluded parts. As shown in \cref{tab:all_corres}, \ours~experiences an average drop of $2.9\%$ PCK@$0.1$ between modal and amodal correspondences, confirming that occlusions pose a greater challenge, yet performance remains competitive.  

\emph{Normalized Object Space.} Finding correspondences using a normalized object space alone is insufficient. We can see this from the NOCS baseline, which \ours~outperforms for both 2D and 3D modal correspondences, and from the fact that \ours~improves over GenPose++, which uses the NOCS space to match query to target points. Qualitatively, in \cref{fig:qualitative}, we observe how GenPose++ incorrectly matches the front of a hairdryer to a location outside the target hairdryer's due to the smaller size.

\emph{MagicPony} is the closest baseline, and its deformations can fit 2D images well in most cases. However, since it is designed for 2D alignment, it struggles to recover consistent 3D rotations across images. As a result, the model tends to compensate through deformation rather than rotation, which, while plausible in 2D, leads to unreliable 3D correspondences (see \cref{fig:qualitative}, bottom-right). This is further evidenced in \cref{tab:all_corres}, where $\text{MagicPony}_{\text{2D}}$ outperforms MagicPony+GP++, confirming that learning entangled canonical and camera space together does not lead to reliable correspondences in 3D. \ours~addresses this through explicit disentanglement of pose, shape, and canonicalization during training, yielding more accurate 3D structure and higher semantic consistency across viewpoints.

\emph{SPair71k.} Thanks to its broader category diversity, our benchmark is more challenging than SPair71k~\cite{Min19SPair}, as seen in the performance gap: DINOv2 achieves $52.7\%$ on SPair71k but drops to only $22.9\%$ on \ourDataset.

\begin{wraptable}{r}{0.43\linewidth}
\vspace{-2.2em}
\caption{\textbf{Real-world evaluation.} PCK@0.1 on a filtered subset of ROPE. \ours~generalizes well.}
\centering
\resizebox{\linewidth}{!}{%
\setlength{\tabcolsep}{5pt}
\begin{tabular}{lcc} 
\cellcolor{gray!20}    & \cellcolor{gray!20}2D@0.1 &\cellcolor{gray!20}3D@0.1 \\
$\text{MagicPony}_{\text{2D}}$        & 16.8                          & N/A \\
GP++          & 37.0                            & 25.1            \\
MagicPony+GP++       & 12.6                          & 7.3\\
\ours           & 44.7                          & 34.8\\
\end{tabular}
}
\label{tab:real_data}
\vspace{-3em}
\end{wraptable}

\emph{Real-world subset.} \ourDataset is synthetic but of high quality: object transparency is modeled for depth, appearances are photorealistic, and annotations are exact by construction. In contrast, the real subset of Omni6DPose (\ie, ROPE) relies on pose tracking, which leads to unreliable annotation. Moreover, ROPE scenes contain only 3--5 objects with low occlusion. For these reasons, we do not consider ROPE of sufficient quality to include in the benchmark. Nevertheless, to verify that our model generalize to real-world data, we evaluate on a filtered subset of ROPE. Concretely, we selected 5 representative classes based on dataset statistics (mean and standard deviation across all methods), obtained the 3D scanned instances directly from the original authors, and \textbf{verified alignment frame by frame}. In total, we evaluate on 5 classes (\ie, $24$ instances, $134$ keypoints). As shown in \cref{tab:real_data}, results are consistent with those on synthetic data: 2D performance is on par, while 3D performance drops by approximately $7\%$, which we attribute to the noisier depth and 3D annotation, rather than a failure of generalization. Overall, this confirms that our model can transfer to real-world data. Further details are provided in \cref{suppsec:real_data}.

\subsection{Limitations and Failure Modes}\label{sec:limitations}
While \ours~enables semantically consistent correspondences across diverse object instances, several limitations remain.
\emph{(i) Topology:} a shared template with fixed connectivity cannot handle large topological variation (\eg, missing parts).
\emph{(ii) Pose sensitivity:} Since correspondences in camera space rely on accurate pose estimation, large pose errors cause global misalignment even when shape deformation is correct. Jointly optimizing pose and deformation remains an open problem.
\emph{(iii) Fine-grained details:} Deformations are regularized to encourage smooth geometry and stable training, which can oversmooth thin structures.
Despite these limitations, our experiments demonstrate that identity-preserving morphable priors provide a highly effective mechanism for establishing monocular category-level 3D correspondences in camera space. More details in \cref{suppsec:additional_results}.

\section{Conclusion}
This paper introduces a paradigm shift from correspondence evaluation in 2D camera space or 3D object space toward \textbf{category-level 3D correspondences in camera space}. 
\textbf{\ourDataset} provides 50 everyday categories in crowded scenes with mesh-based annotations, establishing a solid foundation for comparing monocular 3D correspondence methods with explicit handling of symmetries, occlusions, and challenging amodal correspondences.
We demonstrate that solving this task requires moving beyond 2D feature matching. 
\textbf{\ours} leverages morphable priors to achieve state-of-the-art performance through pose- and occlusion-aware supervision, successfully morphing objects while maintaining consistent correspondences across instances with varying shapes and poses.
We also show that approaches relying only on 2D supervision remain insufficient.
Our benchmark provides a foundation for expanding correspondence learning toward embodied robotics applications, where reasoning about full 3D object geometry, including occluded parts, is essential.

{
    \small
    \newpage
    \section*{Acknowledgments}
AK acknowledges support via his Emmy Noether Research Group funded by the German Research Foundation (DFG) under grant number 468670075.
This research was funded by the Deutsche Forschungsgemeinschaft (DFG, German Research Foundation) under grant number 539134284, through EFRE (FEIH\_2698644) and the state of Baden-Württemberg. 
\begin{center}
\includegraphics[width=0.3\textwidth]{figures/acknowledgement/BaWue_Logo_Standard_rgb_pos.png} ~~~ \includegraphics[width=0.3\textwidth]{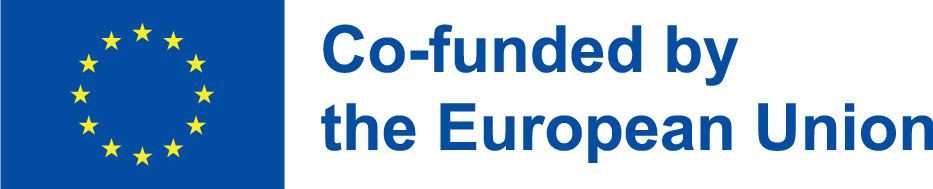} 
\end{center}
    \bibliographystyle{author-kit/splncs04}
    \bibliography{reference}
}

\newpage
\maketitlesupplementary
\appendix

\renewcommand{\thefigure}{A\arabic{figure}}
\renewcommand{\thetable}{A\arabic{table}}
\renewcommand{\theequation}{A\arabic{equation}}
\renewcommand{\theHfigure}{A\arabic{figure}}
\renewcommand{\theHtable}{A\arabic{table}}
\renewcommand{\theHequation}{A\arabic{equation}}
\renewcommand{\theHsection}{A\arabic{section}}
\renewcommand{\theHsubsection}{A\arabic{section}.\arabic{subsection}}
\renewcommand{\theHsubsubsection}{A\arabic{section}.\arabic{subsection}.\arabic{subsubsection}}
\setcounter{figure}{0}
\setcounter{table}{0}
\setcounter{equation}{0}

\newcommand{\additem}[2]{%
\item[\textbf{(\ref{#1})}] 
    \textbf{#2} \dotfill\makebox{\textbf{\pageref{#1}}
    }
}

\newcommand{\myindent}{.5em}
\newcommand{\addsubitem}[2]{%
\vspace{.5em}
    \textbf{(\ref{#1})}
        \hspace{\myindent} #2 \\    
}

\setlist[itemize]{noitemsep,leftmargin=*,topsep=0em}
\setlist[enumerate]{noitemsep,leftmargin=*,topsep=0em}

\noindent This supplementary material provides additional details and results complementing the main paper. We elaborate on the limitations of existing benchmarks, present extended quantitative and qualitative results, and describe our experimental setup and baselines. We further detail the statistics of \ourDataset, discuss the real subset of Omni6DPose, and describe the mesh annotation process. Finally, we provide details on auxiliary training losses, introduce our HueGrid visualization, and discuss correspondence evaluation under symmetry and occlusion.%

\vspace{0.3in}

\begin{enumerate}[label={({\arabic*})}, topsep=1em, itemsep=.2em]
    \additem{suppseclimitation_existing_bench}{Limitation of existing benchmarks}\\[0.4em]\hfill
    
    \additem{suppsec:additional_results}{Additional results}\\[0.4em]\hfill

    \additem{suppsec:baselines}{Experimental details}\\[0.4em]\hfill

    \additem{suppsec:additional_stats}{Additional dataset statistics}\\[0.4em]\hfill
    
    \additem{suppsec:real_data}{Real subset of Omni6DPose}\\[0.4em]\hfill

    \additem{suppsec:annotation}{Mesh annotation process}\\[0.4em]\hfill

    \additem{suppsec:losses}{Additional Losses}\\[0.4em]\hfill

    \additem{suppsec:visualization}{HueGrid Visualization}\\[0.4em]\hfill

    \additem{suppsec:correspondence_evaluation}{Discussion about correspondence evaluation}\\[0.4em]\hfill
\end{enumerate}

\newpage
\section{Limitation of existing benchmarks}
\label{suppseclimitation_existing_bench}

\begin{figure}
  \centering
  \includegraphics[width=0.75\linewidth]{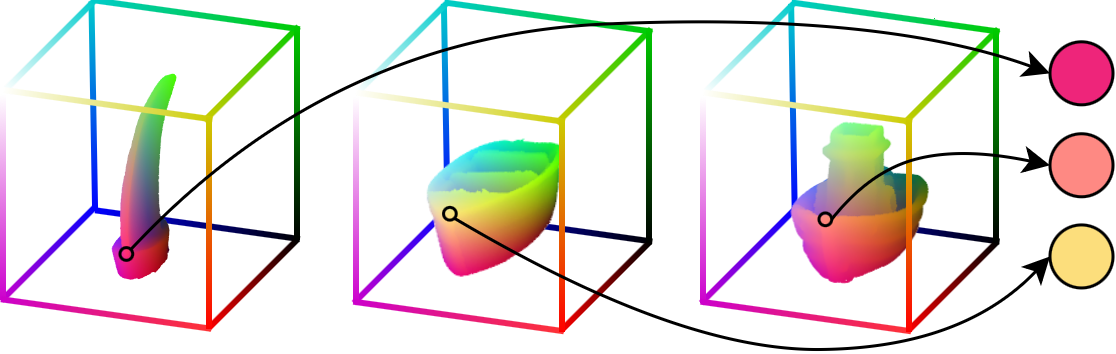}
  \caption{\textbf{NOCS is not semantically consistent.} The same semantic point (\eg, the bow) is marked across three boat instances, yet it maps to different NOCS coordinates in each case. This illustrates that NOCS encodes geometry relative to the geometry rather than semantic part identity, making it an unreliable representation for correspondence evaluation.}
  \label{fig:nocs_issue}
  \vspace{-1em}
\end{figure}

Normalized Object Coordinate Space (NOCS)~\cite{wang2019normalized} maps each visible pixel to a point in a canonical $[0,1]^3$ cube aligned with the object's bounding box. While this encodes a form of 3D information, the representation is purely geometric: coordinates are assigned based on spatial position within the object's bounding box, without any notion of semantic part identity. As a consequence, two points sharing the same NOCS coordinates may correspond to entirely different semantic parts if the geometry of the two instances differs---\eg, the bow of a narrow boat and the bow of a wide one occupy different NOCS locations, while two geometrically similar but semantically distinct regions may coincide (see \cref{fig:nocs_issue}). This means that NOCS-based matching can only succeed when shape variation across instances is small.

Interestingly, as reflected in our results, using NOCS as a feature for 3D matching can still yield reasonable performance in some settings, since geometric proximity is often a adapted proxy for semantic similarity when categories are sufficiently rigid. However, this correlation breaks down for categories with high intra-class shape variation, and critically, NOCS provides no principled way to evaluate whether a predicted correspondence is semantically correct. Using NOCS coordinates as ground truth for correspondence evaluation therefore treats geometric coincidence as semantic alignment, making it an unreliable proxy for the task we seek to evaluate.

\section{Additional results} 
\label{suppsec:additional_results}
In addition to the results reported in the main paper, we provide in \cref{supptab:all_results_2D,supptab:all_results_3D} the complete set of quantitative results for our method, covering all categories of \ourDataset. 
These extended results complement the main text by offering a more fine-grained view of per-class performance. 
Importantly, we observe the same overall trends as in the main paper. 
This consistency arises because the categories highlighted in the main figures were chosen at random, rather than being selected to favor particular outcomes. 
Thus, the additional results confirm that our observations hold uniformly across the entire benchmark and are not biased by the choice of examples shown in the main paper.

\begin{figure}[h]
    \centering
    \includegraphics[width=\linewidth]{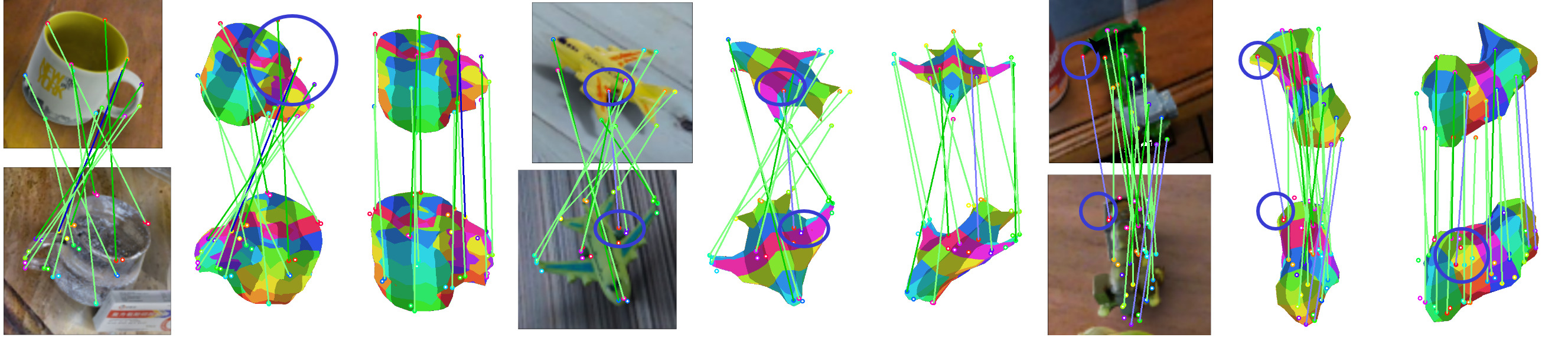}
    \caption{\textbf{Qualitative Results.} We illustrate some limitations qualitatively. In the first example, the pose estimation for the query object is slightly off, resulting in wrong projections on the estimated mesh. Second, coarse estimation of the mesh results in wrong correspondence. Third, wrong depth estimation, leads to wrong 3D correspondence estimation, despite the 2D projection is accurate.}
    \label{fig:suppl_limits}
    \vspace{-1em}
\end{figure}

Despite the overall robustness of our method, some limitations can be observed in challenging scenarios. A first source of error arises from inaccurate pose estimation from \cite{omni6Dpose}. Since canonical alignment is a prerequisite for predicting consistent correspondences, pose misalignment can propagate through the pipeline and lead to incorrect predictions. 
A second limitation concerns the deformation decoder. The learned deformations are constrained by both the template representation and the distribution of training data. As a result, objects that exhibit high intra-class variability, or that contain fine-scale structures not well captured in the template, often cannot be deformed adequately. This is especially evident for thin or elongated extremities such as airplane wings, bottle tips, or animal legs, where the predicted deformation either underestimates the required displacement or, in extremely rare cases, collapses the geometry entirely.  
Finally, the model may fail in cases where very large non-linear deformations are required. Since the decoder is trained to interpolate within the observed shape distribution, extrapolations to unseen structural variations remain difficult. Consequently, regions that extend far beyond the canonical template tend to remain under-deformed, leading to visible artifacts such as truncated parts or floating geometry. While these errors are relatively rare, they underscore the inherent trade-off between enforcing a shared canonical prior and maintaining sufficient flexibility to capture extreme shape variations across object instances. 
We also provide additional qualitative limitations in \cref{fig:suppl_limits}.

\section{Experimental details} 
\label{suppsec:baselines}

\subsection{Hyperparameters} \label{suppsec:exp_hyperparams}
Training \ours involves multiple components and multiple losses, so we draw inspiration from~\cite{Common3D,wu2023magicpony,omni6Dpose} for our hyperparameter settings. \cref{tab:hyperparams} summarizes the overall training setup, loss weights, and model architectures used across our experiments.

\subsection{DINOv2}
\label{suppsec:dinov2D}
For the DINOv2 baseline, we use the ViT-S backbone initialized from the public weights. Images are resized to $448^2$, yielding a $32^2$ patch grid, and we L2-normalize the resulting feature map before computing correspondences.

\subsection{NOCS}
\label{suppsec:NOCS}
We closely follow the procedure introduced by \cite{wang2019normalized} to evaluate the NOCS baseline on \ourDataset. We use the same ResNet50~\cite{resnet} backbone together with Feature Pyramid Network (FPN). For every training image we generate ground-truth NOCS targets by normalizing each object mesh to the unit cube and encoding the resulting XYZ coordinates directly as RGB values. Using the camera poses provided in Omni6DPose\cite{omni6Dpose}, we then render these NOCS maps so that every pixel stores its canonical 3D coordinate. Training uses the official ground-truth instance masks, category labels, and depth maps from Omni6DPose to supervise the model and to restrict supervision to the visible object regions.
At inference time we predict a dense NOCS map for each input image. 
For 2D correspondence queries, we read the predicted canonical coordinate at the query pixel and find the nearest neighbor in NOCS space among all image pixels in the target image; the location of that neighbor serves as the correspondence prediction.
For 3D correspondence queries, given a 3D query point $\xquery$ in the source image, we first find the corresponding canonical coordinate by projecting $\xquery$ into the source image and reading the predicted NOCS value at that pixel. We then find the nearest neighbor in NOCS space among all pixels in the target image; we back-project that pixel using the depth map to obtain the predicted 3D correspondence $\xtarget$.

\begin{table*}[t]
\centering
\footnotesize
\setlength{\tabcolsep}{6pt}
\renewcommand{\arraystretch}{1.2}
\begin{tabular}{@{} p{0.25\linewidth} l @{\hspace{1em}} p{0.35\linewidth} l @{}}
\toprule
\multicolumn{2}{c}{\textbf{Training Hyperparameters}} & \multicolumn{2}{c}{\textbf{Loss Weights}} \\
\cmidrule(r){1-2}\cmidrule(l){3-4}
Optimizer & Adam & Mesh Chamfer Distance ($\glossrec$) & 0.1 \\
Batch Size & 30 & Mask Mean Square Error ($\glossmask$) & 2 \\
Batch Accumulation & 2 & Mask Dist. Transform ($\glossmaskdt$) & 200 \\
Learning Rate & $1.0\times 10^{-3}$ & SDF Regularization ($\glosseik$) & 0.01 \\
Epsilon & $1.0\times 10^{-8}$ & Deformation Regul. ($\glossregdef$) & 0.075 \\
Beta 1 & 0.9 & Smoothness Regul. ($\glossregdefsm$) & 0.0075 \\
\cmidrule(l){3-4}
Beta 2 & 0.999 & \multicolumn{2}{c}{\textbf{Template Architecture}} \\
\cmidrule(l){3-4}
Weight Decay & 0 & Type & Coord. MLP \\
LR Scheduler & Exponential LR & Layers & 5 \\
Warmup & 100 & Hidden Dimension & 256 \\
Gamma & 0.98 & Out Dimension & 1 \\
LR Min. & $1.0\times 10^{-4}$ & DMTet Resolution & 16 \\
\midrule
\multicolumn{4}{c}{\textbf{Deformation Architecture}} \\
\midrule
Backbone & DINOv2 ViT-S & Deformation Decoder & Coord. MLP \\
Deformation Encoder & ResNet Blocks & Layers & 5 \\
ResNet Blocks & 4 & Hidden Dimension & 256 \\
ResNet Block Type & bottleneck & Out Dimension & 6 \\
Out Dimensions & ${[256]}^4$ & & \\
Strides & {[2, 2, 2, 2]} & & \\
Pre-Upsampling & {[1, 1, 1]} & & \\
\bottomrule
\end{tabular}
\vspace{2em}
\caption{Full-width hyperparameter overview including training setup, loss weights, and model architectures.}%
\label{tab:hyperparams}
\end{table*}

\subsection{MagicPony}
\label{suppsec:magic_pony}
Following MagicPony~\cite{wu2023magicpony}, we sample 5K images, extract object features using the provided modal masks, and apply PCA to reduce the feature dimension to 16. We replace the original DINOv1 encoder with DINOv2, which improves category-level 2D correspondence estimation~\cite{zhang2023tale}. Due to memory constraints, each category-level model is trained for 120 epochs with a grid resolution of 128, whereas the original implementation switches to resolution 256 for the final 30 epochs. Because our evaluation emphasizes correspondence accuracy within a $10\%$ object-size tolerance rather than fine-grained reconstruction, sub-percent shifts (e.g., $<0.5\%$) are negligible.

\section{Additional dataset statistics} 
\label{suppsec:additional_stats}

We rely exclusively on the realistic synthetic subset of Omni6DPose \cite{omni6Dpose}. 
Preliminary experiments showed that the real captures provide limited diversity: most categories contain at most two unique object instances, scenes are often repeated across long video sequences, and overall variation in layout is low. 
As a result, the number of reliable correspondences that can be established from the real subset is severely restricted.  

\begin{figure}[ht]
    \centering
    \includegraphics[width=\linewidth]{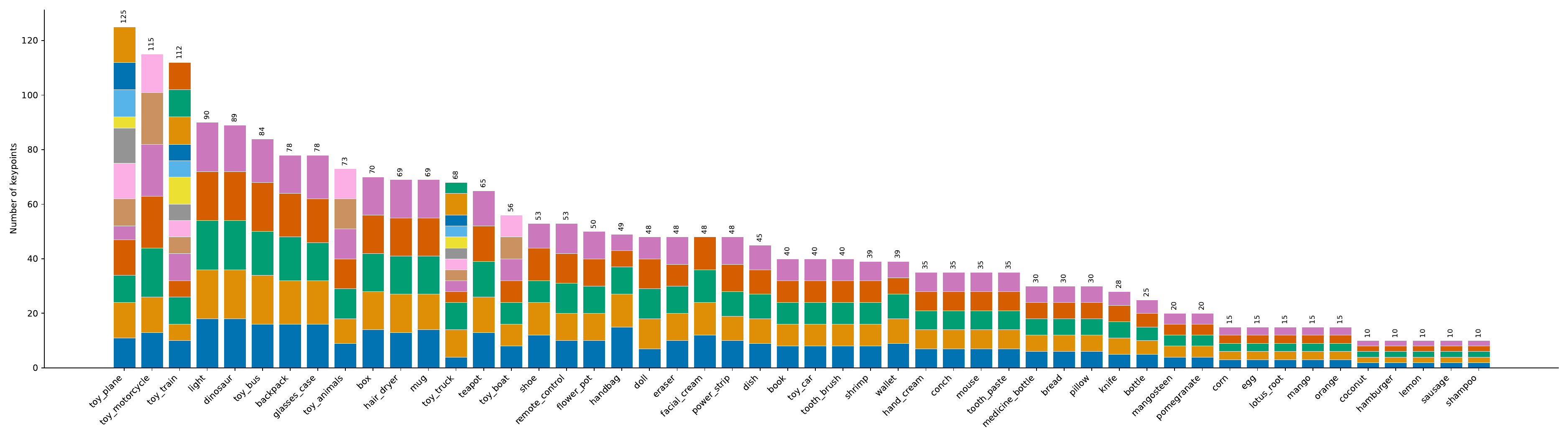}
    \caption{
    \textbf{Total number of annotated keypoints per class.} Different object instances are shown in different colors. \emph{Note.} The number of keypoints per instance can vary within a class because instances often differ in shape and semantics. For example, two \texttt{toy\_plane} instances have fewer keypoints because they are helicopters, and roughly half of the \texttt{toy\_train} instances are high-speed bullet trains while the others are conventional locomotives.}
    \label{suppfig:num_kp_per_class}
\end{figure}

In contrast, the synthetic pipeline offers large-scale variation in both object instances and scene composition. 
This diversity is crucial for learning robust 2D--3D semantic correspondences across categories. 
Moreover, the synthetic subset has been designed to closely mimic real-world conditions, with natural lighting, cluttered environments, and realistic occlusions. 
This ensures that models trained on our benchmark generalize well beyond simplified synthetic settings.  
Therefore, our benchmark focuses on the high-quality synthetic subset, which provides both realism and sufficient coverage for large-scale correspondence evaluation. 
In total, \ourDataset~contains 178k images across 280 unique object instances from 50 categories, making it the first large-scale dataset with dense, semantically consistent 2D--3D correspondences for everyday objects.
To better illustrate the scope of the annotations, \cref{supptab:nb_keypoints} reports the number of annotated keypoints for each category, highlighting differences in semantic coverage across classes. In \cref{suppfig:num_kp_per_class}, we further visualize the total number of keypoints annotated per class and indicate, through color coding, how many object instances were annotated. Together, these results offer a clear overview of the dataset’s scale and diversity and underscore its suitability as a benchmark for category-level 3D correspondence.

\section{Real subset of Omni6DPose}
\label{suppsec:real_data}
To evaluate on real data, we select a representative subset of classes from Omni6DPose.
Our goal is to identify a small set of classes that faithfully reflects the statistical properties of the full benchmark, while minimizing annotation and evaluation effort.
\begin{figure}[t]
  \centering
  \begin{subfigure}[b]{0.48\linewidth}
    \includegraphics[width=\linewidth]{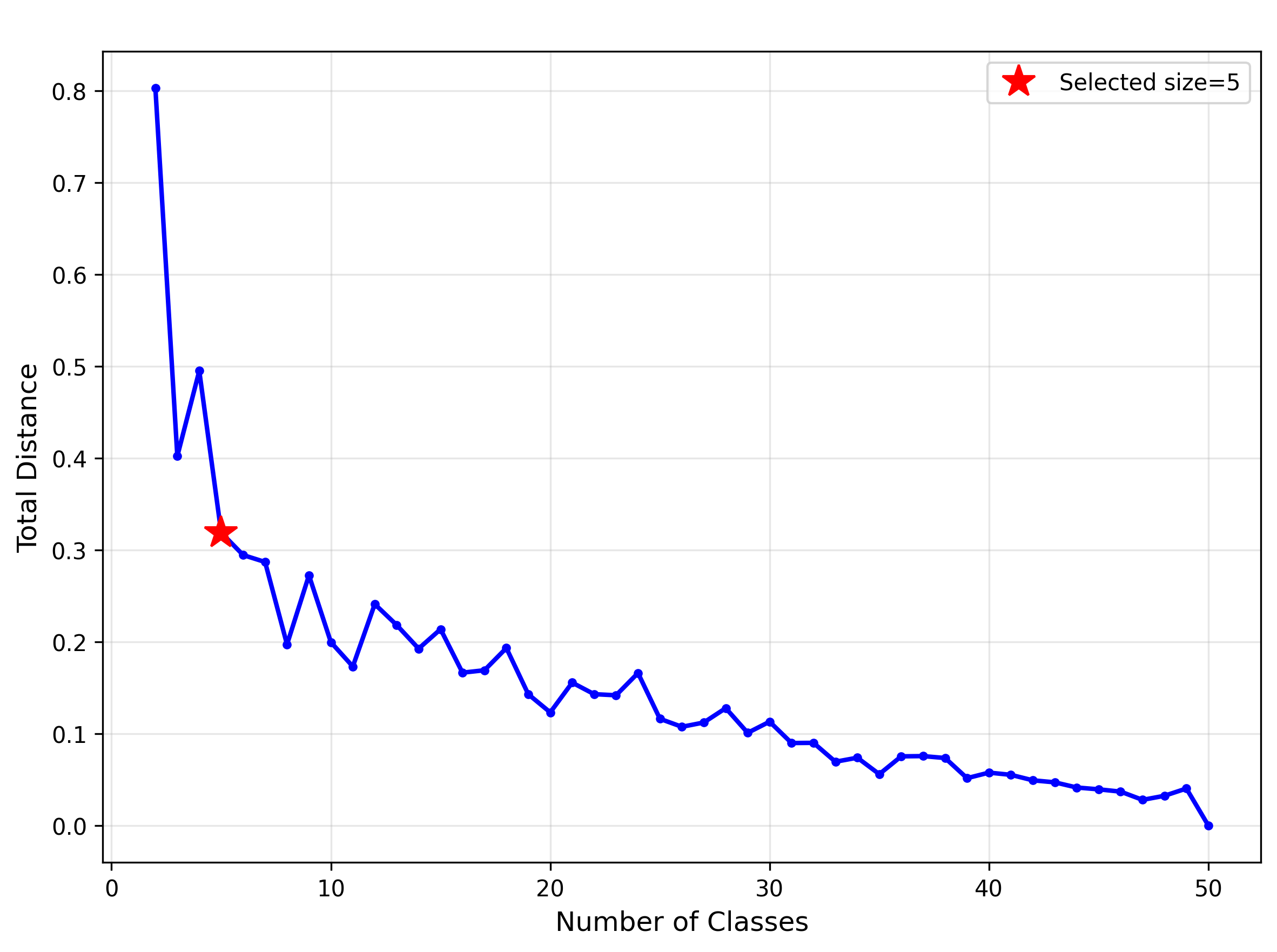}
    \caption{Total distance to full-dataset statistics as a function of subset size $k$.}
    \label{suppfig:optimal_size_total}
  \end{subfigure}
  \hfill
  \begin{subfigure}[b]{0.48\linewidth}
    \includegraphics[width=\linewidth]{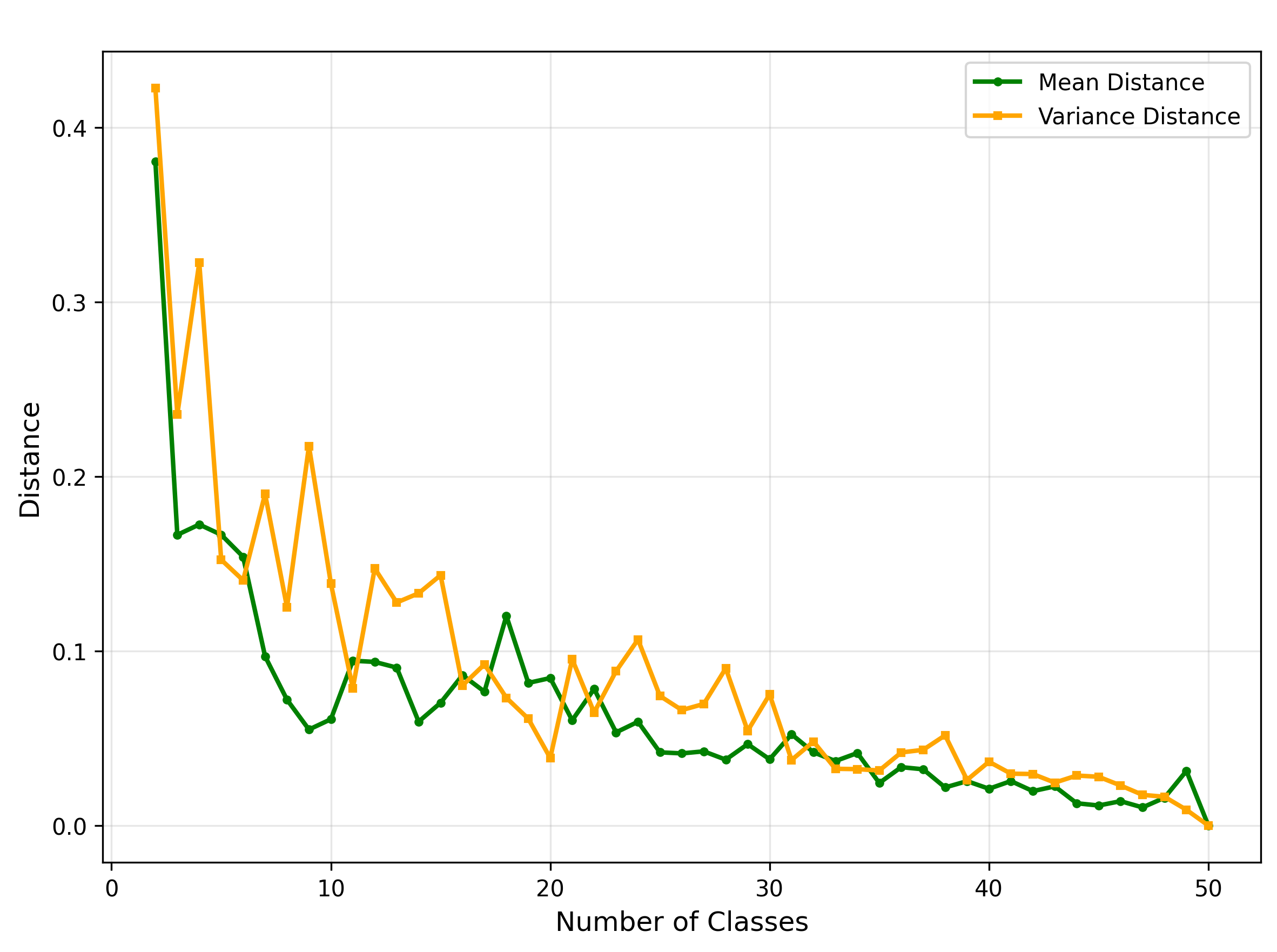}
    \caption{Decomposition into mean and variance distance components.}
    \label{suppfig:optimal_size_components}
  \end{subfigure}
  \caption{Optimal subset size analysis.
  The distance decreases rapidly with $k$ but plateaus beyond $k{=}5$, which we select as the optimal trade-off between representativeness and evaluation cost.}
\end{figure}

\paragraph{Optimal number of classes.}
We first determine the ideal subset size by testing all sizes from 2 to 50 classes.
For each size $k$, we sample $10'000$ random unique subsets of classes and measure how well their statistics match the full dataset and select the minimum distance across all subsets of that size.
Concretely, for each subset we compute, per method, the mean and variance of PCK@0.1 scores across the selected classes (using 3D Modal results), then measure the normalized Euclidean distance to the corresponding statistics computed over all 50 classes.
As shown in \cref{suppfig:optimal_size_total,suppfig:optimal_size_components}, the total distance decreases rapidly as $k$ increases, but the marginal gain flattens beyond $k{=}5$.
We therefore select $k{=}5$ as the optimal subset size, balancing representativeness and annotation cost.

\paragraph{Optimal class selection.}
Given $k{=}5$, we exhaustively search over all $\binom{50}{5}$ combinations to find the subset minimizing the total distance (sum of mean-distance and variance-distance across all evaluated models).
The selected 5 classes are: \emph{bread}, \emph{facial cream}, \emph{hair dryer}, \emph{handbag}, and \emph{tooth brush}.
This subset achieves a total distance of $0.116$ to the full-dataset statistics (mean distance: $0.109$, variance distance: $0.123$).

\paragraph{Real subset annotation and evaluation.}

\begin{figure}
    \centering
    
    \includegraphics[width=1.\linewidth]{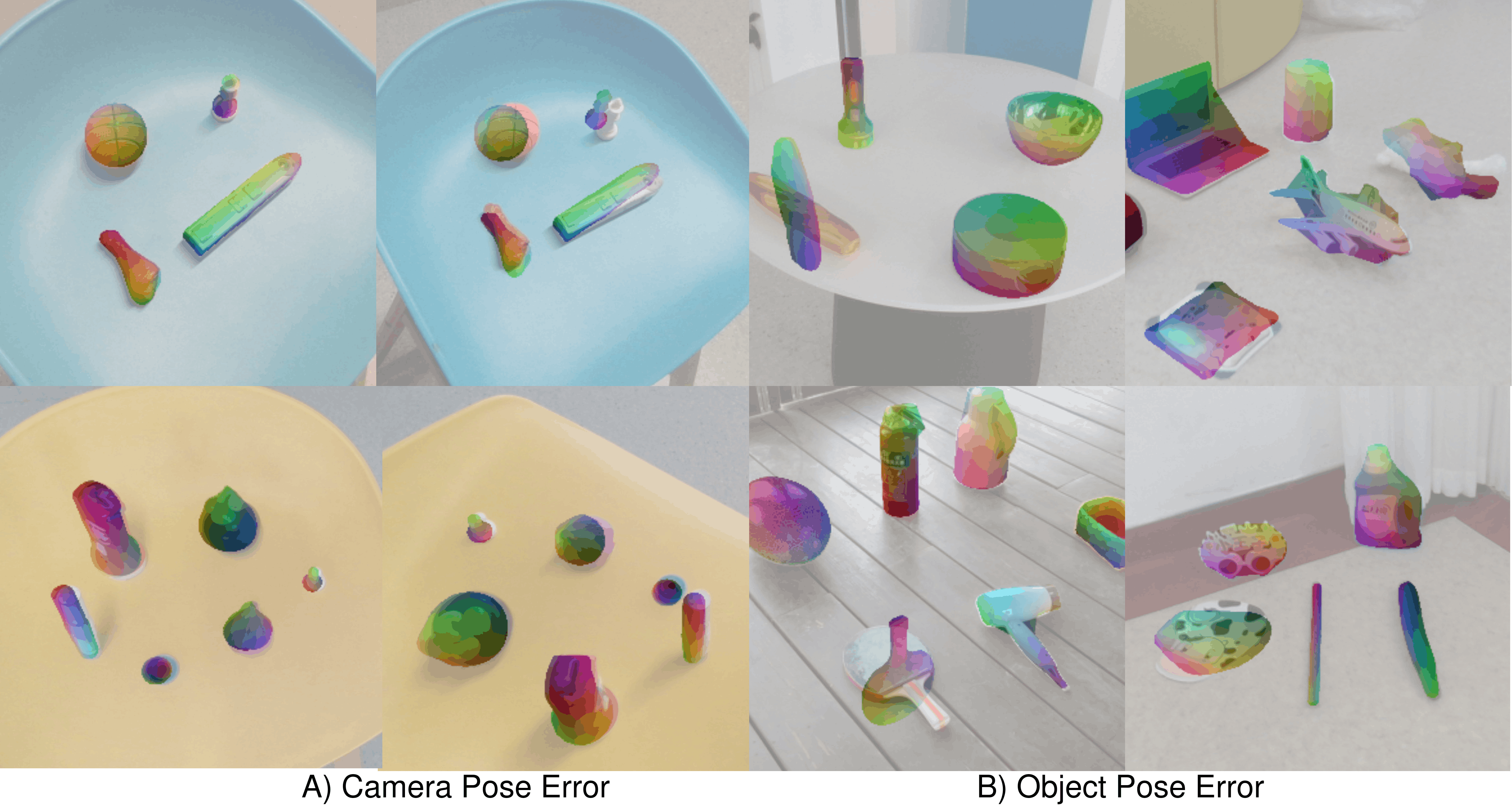} 
    \caption{\textbf{Alignment Errors in the Omni6DPose Real Subset.}
    We overlay RGB images with projections of approximate object meshes to visualize transformation errors in the real subset of Omni6DPose. Since 3D keypoints are annotated in canonical space and subsequently transformed into camera space, incorrect transformations lead to misaligned keypoints. We observe two primary error sources. \textit{(A) Camera pose errors:} tracking failures cause incorrect camera poses in some frames. \textit{(B) Object pose errors:} inaccurate object tracking or misalignment between pose annotations and the (unpublished) object meshes leads to incorrect projections. In our curated subset of the real dataset, we remove approximately $13\%$ of the images exhibiting such issues, and evaluate exclusively on the remaining correctly annotated frames.}
    \label{suppfig:real_alignment_issue}
\end{figure}

We obtained the 3D scanned instances directly from the original authors of Omni6DPose.
Since the real subset consists of multi-frame videos where the camera moves around a static scene, we needed to \textbf{verify the 3D annotation alignment frame by frame}.
To do so, we reprojected the 3D assets into camera space and checked consistency between the projected mesh and the RGB image throughout each video.
We found that the alignment was accurate in some frames but drifted across the sequence, indicating inaccuracies in the provided 3D poses, as illustrated in \cref{suppfig:real_alignment_issue}.
We therefore manually curated the data by retaining only frames where the 3D annotation was visually consistent with the RGB image, resulting in filtering out $13\%$ of all frames.
In parallel, we followed the same keypoint annotation process described in \cref{suppsec:annotation} to label 3D keypoints on the real instances.
This yielded $24$ instances and $134$ keypoints across 5 classes.
We then evaluated all models from the main paper on this real subset using the same protocol and metrics.

\section{Mesh annotation process}
\label{suppsec:annotation}

For mesh annotation, we convert each CAD mesh into a point cloud to facilitate visual inspection and interaction. 
Annotators are then provided with up to 20 3D keypoints per category that must be placed consistently across all instances. 
These keypoints are chosen to be semantically meaningful and geometrically well-defined: rather than marking the center of a continuous surface, annotators focus on distinctive structures such as corners, edges, wheel centers, handles, or wing tips. 
This strategy ensures that annotated points are both discriminative and reliably transferable across different instances of a category.  
To guarantee annotation quality, each instance was independently annotated by two annotators. 
The two annotation sets are then automatically merged using a correspondence-based algorithm. 
First, keypoints from both annotators are transformed to an object-centric coordinate frame and mutual nearest-neighbor correspondences are computed across all instances of a category. 
Matched keypoints are either classified as close (within $5\%$ of the object's bounding-box diagonal) or distant. 
Based on matching patterns across instances, keypoints are automatically accepted (pairs are always matched and close, \texttt{AUTO\_ACCEPT}), split into separate entries (pairs are always matched but distant, indicating semantic disagreement between annotators, \texttt{AUTO\_SPLIT}), or kept as-is (never matched, \texttt{AUTO\_UNMATCHED}). 
Ambiguous cases (\ie, all remaining keypoints not falling in any previous categories), which includes keypoints with inconsistent matching behavior or mixed proximity patterns, are resolved through an interactive post-merging step, where both annotators visualize correspondences across multiple instances and manually and mutually decide whether to accept a single keypoint (\eg, \texttt{MANUAL\_ACCEPT} + \texttt{SET1} \& \texttt{REJECT} for the second keypoint), merge keypoints (\ie, use the mean of both keypoints, \texttt{MANUAL\_ACCEPT} + \texttt{MEAN}), split keypoints (\ie, create separate keypoints and keep both when they refer to different semantic concepts, \texttt{MANUAL\_ACCEPT} + \texttt{SET1}\&\texttt{SET2}), or reject both keypoint (\texttt{REJECT}). We summarize the final merged status and manual decision distributions in \cref{suptab:merged_status}.
This systematic merging procedure reduces noise and ensures high-quality annotations consistent across the dataset.
In addition, the reference mesh for each category was annotated first, and subsequent instances were aligned to this reference using a 3D interface. 
This alignment step further reduced ambiguities and ensured that annotations across different instances adhered to the same semantic standard. 
Overall, this process yields a compact yet semantically robust set of 3D keypoints that serve as the foundation for our correspondence benchmark.

\begin{table}[h]
    \centering
    \footnotesize
    \caption{Final merged status (left) and manually accepted decision (right) distributions over all categories.}\label{suptab:merged_status}
    \vspace{-.5em}
    \begin{tabular}{lr}
        \toprule
        Status & Percentage \\
        \midrule
        \texttt{AUTO\_ACCEPT} &  23.1\% \\
        \texttt{AUTO\_SPLIT} &  16.6\% \\
        \texttt{AUTO\_UNMATCHED} &  24.9\% \\
        \texttt{MANUAL\_ACCEPT} & 21.9\% \\
        \texttt{REJECT} & 13.4\% \\
        \bottomrule
    \end{tabular}
    \hspace{5em}
    \begin{tabular}{lr}
        \toprule
        Decision & Percentage \\
        \midrule
        \texttt{MEAN}  & 48.2\% \\
        \texttt{SET1}  & 24.5\% \\
        \texttt{SET2}  & 27.3\% \\
        \bottomrule
    \end{tabular}
    
    \vspace{0.4em}
    \parbox{0.9\linewidth}{\emph{Note.} 
    Most rejected keypoints occur when only one annotator set (\texttt{SET1} or \texttt{SET2}) is retained during manual validation. This happens when both sets target the same semantic zones, but one of them is judged to be of relative higher quality and the other is therefore discarded.}

\end{table}

\label{suppsec:mesh_annot}
\begin{figure}[ht]
    \centering
    \begin{subfigure}[t]{\linewidth}
        \centering
        \includegraphics[width=\linewidth]{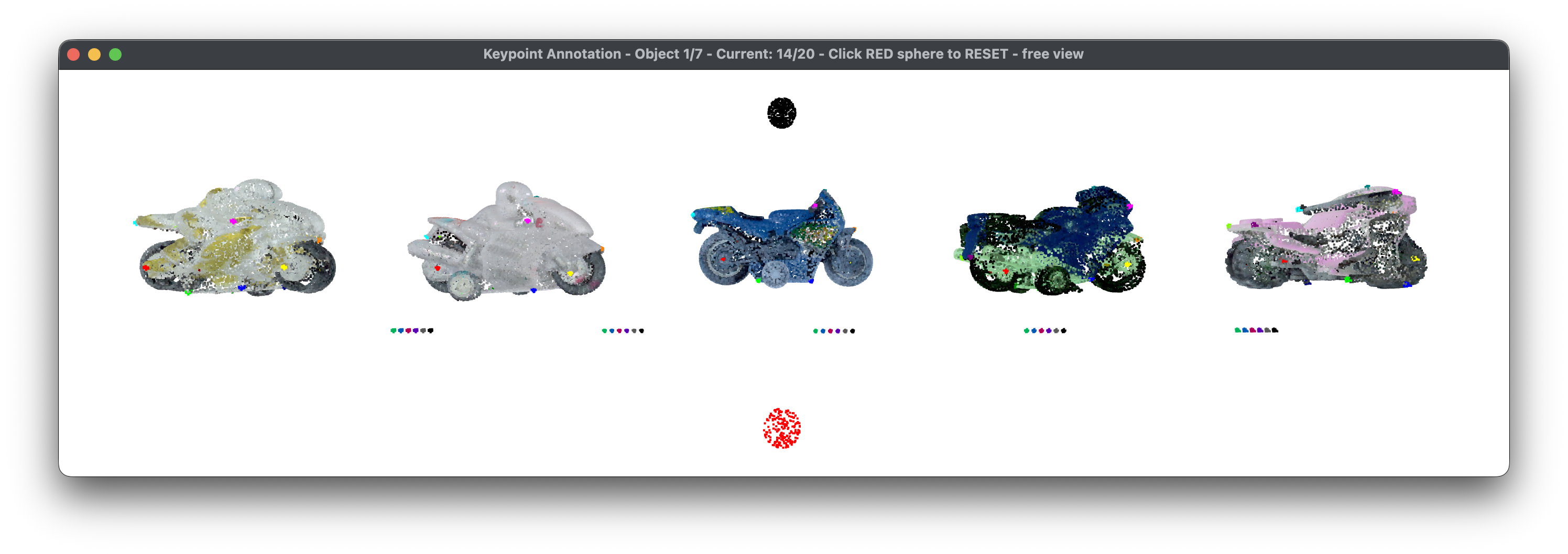}
        \caption{\textbf{Overview of the annotation tool.} Annotated keypoints are displayed directly on the point cloud, allowing annotators to verify their placement.}
        \label{suppfig:annotation-1}
    \end{subfigure}
    \hfill
    \begin{subfigure}[t]{\linewidth}
        \centering
        \includegraphics[width=\linewidth]{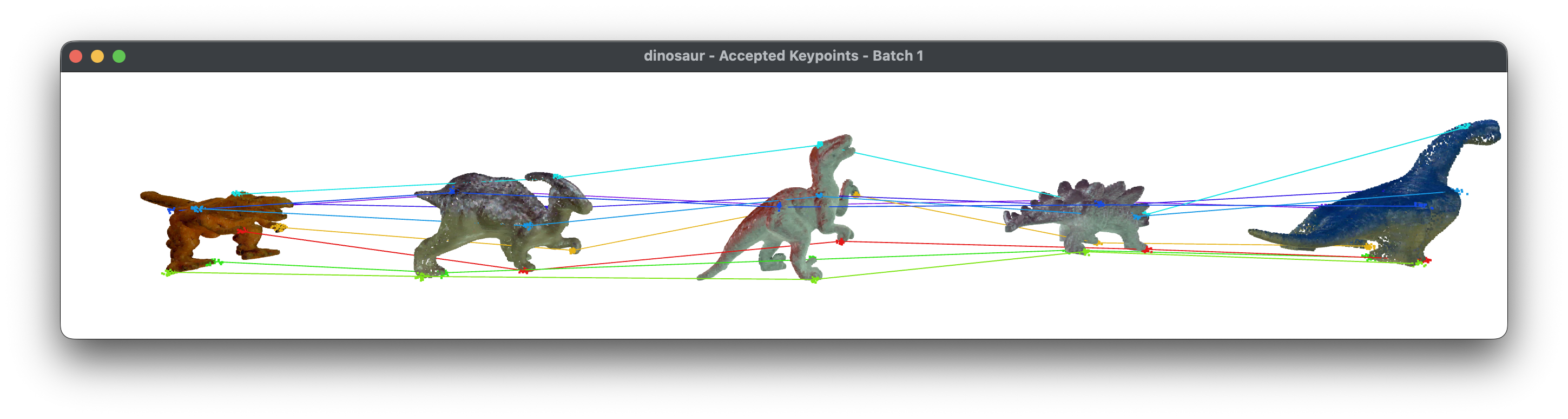}
        \caption{All annotated keypoints and their correspondences for the dinosaur category are visualized, enabling inspection of annotation quality.
}
        \label{suppfig:annotation-2}
    \end{subfigure}
    \caption{\textbf{Annotation process illustration.} Using our interactive 3D interface, annotators align 5 instances per category and assign 3D keypoints to their respective meshes. We also visualize the resulting correspondences to assess their quality and consistency.}
    \label{suppfig:annotation}
\end{figure}

\section{Additional Losses}
\label{suppsec:losses}
Learning accurate correspondences requires not only supervision on visible matches but also strong geometric regularization to stabilize training and enforce plausible shapes.  
To this end, we use additional loss terms (\cref{supp:eq:1,supp:eq:2,supp:eq:3}) that impose additional constraints to the learned deformation and shape representation.

\paragraph{Eikonal loss.}  
To enforce the signed distance function (SDF) property, we adopt the Eikonal regularizer~\cite{gropp2020implicit}, which encourages unit-norm gradients of the implicit function.  
Because gradients are only reliable near the extracted surface, we additionally sample auxiliary points $\pointssdf$ throughout the canonical space:
\begin{equation}\label{supp:eq:1}
    \losseik(\mesh, x) = \big(\lVert \nabla \sdff(x) \rVert_2 - 1\big)^2, 
    \quad x \in \pointssdf.
\end{equation}
This prevents degenerate fields and stabilizes the geometry across unseen regions.

\paragraph{Deformation regularizer.}  
To avoid arbitrary or excessive deformations, we penalize $\ell_2$ deviations of vertices from the category template: 
\begin{equation}\label{supp:eq:2}
    \lossregdef(\mesh, \meshdef, \imagesymbol)= 
    \frac{1}{|\verts|}\sum_{\vv \in \verts } 
    \big\lVert \vv - \affinef(\vv, \instlatent) \big\rVert^2, ~~ \text{with} 
    ~~ \instlatent = \featencl(\imagesymbol)
\end{equation}
This term encourages learned shapes to remain close to the canonical prototype while still allowing instance-specific variation.

\paragraph{Smoothness regularizer.}  
Finally, we promote locally coherent deformations by enforcing smooth displacements across neighboring vertices, following~\cite{zheng2021deep}:  
\begin{equation}\label{supp:eq:3}
\lossregdefsm(\mesh, \meshdef, \imagesymbol) = 
\frac{1}{|\edges|} \sum_{\vi, \vj \in \edges}  
\frac{\big\lVert [\vi - \affinef(\vi, \featencl(\imagesymbol))] 
- [\vj - \affinef(\vj, \featencl(\imagesymbol))] \big\rVert_2}{\lVert \vi - \vj \rVert_2}.
\end{equation}
This regularizer suppresses spurious local distortions while still allowing non-rigid articulation.

Together, these terms ensure that the learned representation respects the SDF property, stays anchored to a canonical template, and maintains smooth, realistic deformations.

\section{HueGrid Visualization} \label{suppsec:visualization}
\begin{figure}
    \centering
    \includegraphics[width=\linewidth]{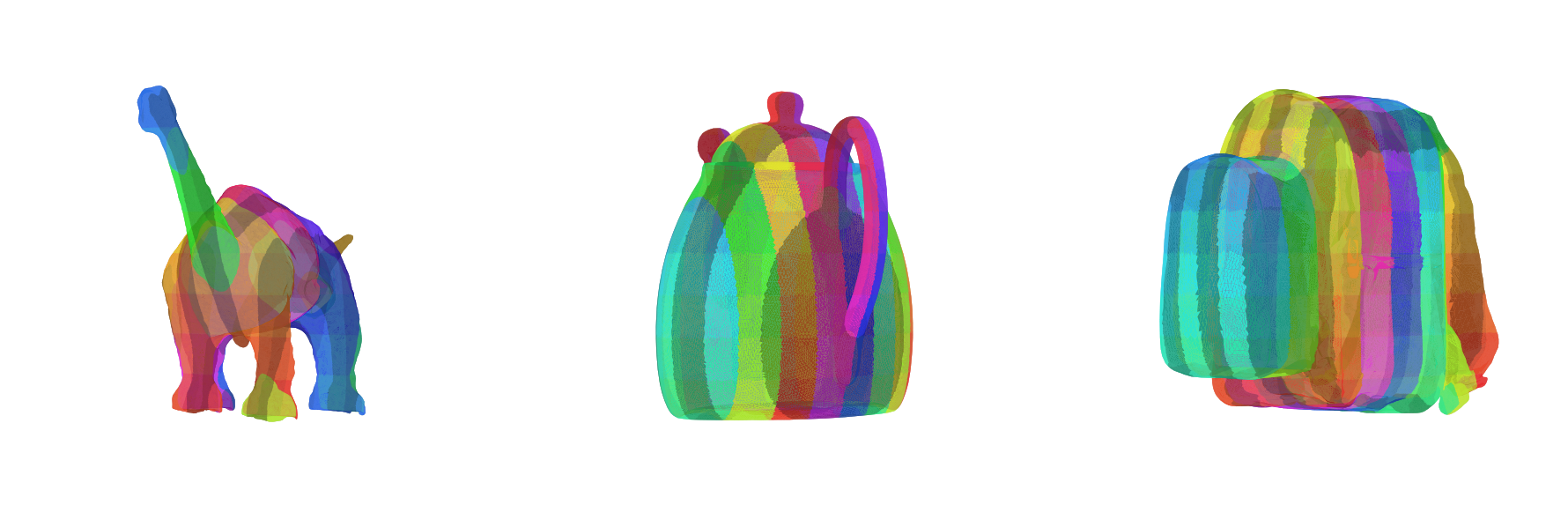}
    \caption{\textbf{HueGrid visualization.} We integrate 3D-based color encoding with a structured checkerboard pattern which allows to jointly highlight absolute correspondences and local deformations. We show the HueGrid projection for three example objects.}
    \label{fig:suppl_viz_rpz}
    \vspace{-1em}
\end{figure}

To visualize dense correspondences, we introduce the \emph{HueGrid} representation. Classical 3D-aware coloring schemes such as NOCS~\cite{wang2019normalized} (widely adopted in~\cite{SHIC,Neverova20,zhu2024densematcher,MeshUp}) encode XYZ coordinates directly as RGB values, but this makes local distortions hard to perceive given the continuous nature of the color mapping. Conversely, \cite{SHIC} texture meshes with a colored checkerboard pattern, which clearly reveals local stretching because square cells deform into visible shapes once projected into the image.

HueGrid combines the best of both ideas: we keep the informative 3D-based color coding of NOCS while superimposing the structured checkerboard cues from~\cite{SHIC}. The resulting visualization simultaneously conveys absolute correspondence information and local geometric deformation. The visualization is illustrated in \cref{fig:suppl_viz_rpz} for three representative mesh examples. We will also provide the code to generate HueGrid visualizations for all meshes and point clouds.

\section{Discussion about correspondence evaluation}
\label{suppsec:correspondence_evaluation}
\begin{figure}
    \centering
    \includegraphics[width=0.35\linewidth]{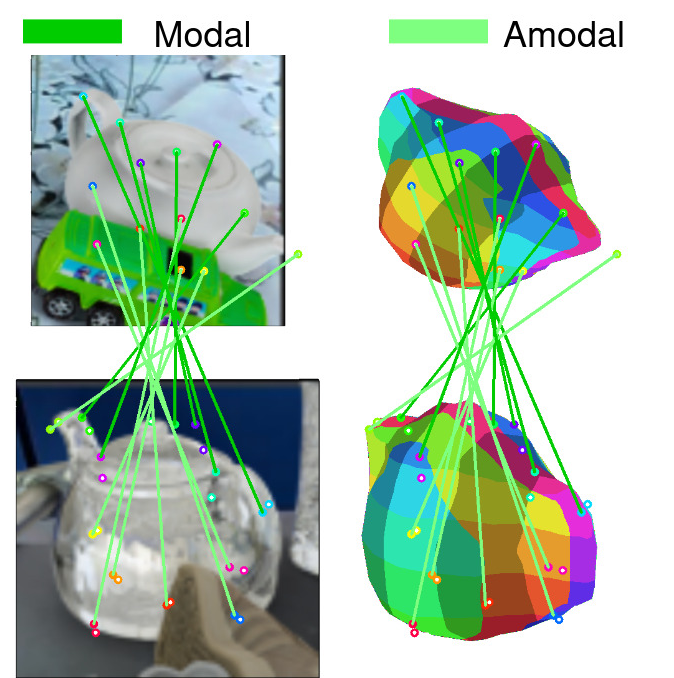}
    \caption{\textbf{Modal vs. Amodal Correspondences.} Choosing the 3D camera space as evaluation space, means we can also evaluate amodal correspondences. Here we show the three types of amodal correspondences in lightgreen, a) self-occlusion, b) occlusion from another object, and c) outside of the camera frustum. Not that it is sufficient if a point is occluded in either the query or the target space. }
    \label{fig:suppl_modal_vs_amodal}
    \vspace{-1em}
\end{figure}

\textbf{Modal vs. Amodal masks.} We distinguish between \emph{modal} and \emph{amodal} correspondences in 3D, see \cref{fig:suppl_modal_vs_amodal}. Modal correspondences are defined only on the subset of surface points that are visible from a given viewpoint, mapping observed 2D pixels to their canonical surface counterparts.  
In contrast, amodal correspondences extend this mapping to the full object surface, including parts that may be (self-)occluded.  
Modal evaluation reflects how well a method can align observed geometry with a canonical template and is directly comparable to tasks such as 2D keypoint transfer.  
Amodal evaluation goes further: it measures whether a model has learned a complete category-level shape prior that can predict correspondences even for unobserved surfaces.  
This distinction is critical for downstream tasks that require holistic understanding, such as shape completion, scene reasoning, or part-level manipulation.  
In 2D, we are restricted to image pixels, which by definition correspond only to visible regions; there is no ground-truth notion of a pixel for an occluded surface.  
In 3D, however, we can explicitly represent the canonical surface $\mathcal{C}$ and predict both visible and occluded points across poses.  
This makes it possible to evaluate amodal correspondences, providing a stronger test of a model’s ability to infer complete, semantically consistent shapes across instances. 

\begin{figure}
    \centering
    \includegraphics[width=\linewidth]{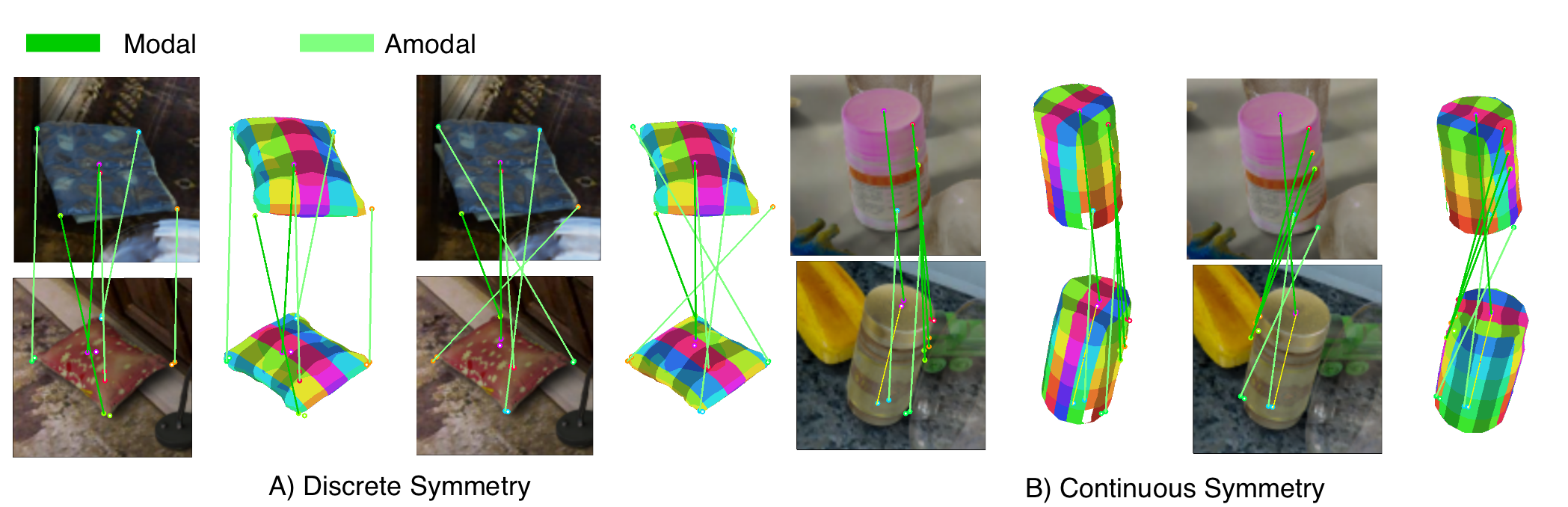}
    \caption{\textbf{Evaluation under symmetry.} Illustration of two correct correspondence estimations with explicit symmety. First, in A), we show two possible set of correct predictions under discrete symmetry. Despite flipping the pillow, the correspondences are correct. Second, in B), we show two possible set of correct predictions under rotational symmetry. We visualize the rotation axis in yellow.}
    \label{fig:suppl_symmetry}
    \vspace{-1em}
\end{figure}

\textbf{Evaluation under symmetry.} Many everyday objects exhibit geometric symmetries that introduce fundamental ambiguities in correspondence. To the best of our knowledge, existing semantic correspondence benchmarks have not addressed symmetries, as they operate purely in 2D where such geometric constraints are difficult to define.
By leveraging 3D annotations, \ourDataset~explicitly handles \emph{discrete} and \emph{continuous} symmetries, ensuring that geometrically equivalent predictions are not unfairly penalized, see \cref{fig:suppl_symmetry}. 

Symmetry is defined as invariance under rotations about a fixed axis. Given a predicted point $\hat{\vx}$ and rotation $R_{\mathbf{a}}(\theta)$ about axis $\mathbf{a}$, the correspondence error is  
\[e_{\text{sym}}(\vx,\hat{\vx}) = \min_{\theta} \|R_{\mathbf{a}}(\theta)\,\vx - \hat{\vx}\|,\]
with $\theta \in [0,2\pi)$ (or specifically $\theta \in \{2\pi k/N\}_{k=0}^{N-1}$ for discrete symmetry with $N$ the number of discrete symmetries).  
Geometrically, this equals the distance from $\hat{\vx}$ to the circular orbit of $\vx$ around the symmetry axis.
With this symmetry-aware definitions, predictions are correct if they align with any symmetric equivalent point.
This yields a fair metric that respects the inherent geometric ambiguities in real-world objects and enables robust evaluation of category-level correspondence methods.

\begin{table}[t]
\footnotesize
\centering
\caption{\textbf{PCK@0.1 results} for 3D, 3D modal, and 3D amodal category-level correspondences comparison across all 50 classes for \ours and the baselines. Beyond showcasing which categories are almost solved versus still challenging, the table reveals how object variability drives performance: classes with low deformation and consistent shapes (\eg, \emph{shampoo}, \emph{corn}) are nearly saturated, whereas highly diverse toy categories (\eg, \emph{toy car}, \emph{toy animal}) remain difficult.}
\vspace{-1em}
\resizebox{\linewidth}{!}{
\begin{tabular}{llllllllllllllllll}
\toprule
 & mean & backpack & book & bottle & box & bread & coconut & conch & corn & dinosaur & dish & doll & egg & eraser & \makecell{facial\\cream} & \makecell{flower\\pot} & \makecell{glasses\\case} \\
\hline
\multicolumn{18}{c}{\cellcolor{gray!20}\textbf{3D}} \\
\hline
GenPose++ (GP++) & 34.3 & 18.8 & 36.9 & 62.8 & 11.1 & 27.5 & 77.0 & 19.6 & 89.8 & 7.8 & 31.8 & 2.8 & 19.5 & 28.0 & 27.4 & 28.0 & 57.2 \\
MagicPony+GP++ & 7.1 & 1.2 & 1.0 & 30.7 & 4.6 & 10.8 & 29.8 & 1.3 & 18.4 & 1.8 & 16.3 & 2.8 & 6.7 & 5.8 & 14.8 & 13.0 & 4.2 \\
\ours & 41.5 & 23.7 & 42.5 & 73.2 & 42.9 & 39.2 & 85.3 & 26.3 & 91.2 & 6.9 & 46.7 & 5.0 & 22.1 & 34.1 & 46.2 & 39.7 & 68.5 \\
\ours w/o Def. & 38.4 & 22.7 & 42.6 & 71.0 & 10.4 & 35.0 & 85.1 & 26.0 & 91.6 & 8.4 & 42.4 & 3.5 & 22.1 & 35.1 & 28.0 & 38.9 & 67.8 \\
\hline
\multicolumn{18}{c}{\cellcolor{gray!20}\textbf{3D Modal}} \\
\hline
DINOv2+D & 24.4 & 5.7 & 18.0 & 29.0 & 27.0 & 16.4 & 52.9 & 16.7 & 40.8 & 8.1 & 15.3 & 2.6 & 10.4 & 39.6 & 31.3 & 28.7 & 37.4 \\
$\text{MagicPony}_{\text{2D}}\text{+D}$  & 14.0 & 3.9 & 4.8 & 32.2 & 23.1 & 22.7 & 27.2 & 10.6 & 24.0 & 4.7 & 10.9 & 0.0 & 13.9 & 20.8 & 26.3 & 21.3 & 26.6 \\
NOCS+D & 26.4 & 6.5 & 31.9 & 58.7 & 42.3 & 24.0 & 69.9 & 2.7 & 75.2 & 6.5 & 34.9 & 1.9 & 6.6 & 18.9 & 26.4 & 22.5 & 51.2 \\
\doublehline
GenPose++ (GP++) & 37.0 & 22.9 & 43.2 & 60.0 & 14.6 & 27.9 & 84.6 & 14.3 & 92.5 & 11.6 & 30.4 & 2.6 & 19.8 & 30.2 & 39.3 & 33.9 & 59.4 \\
MagicPony+GP++ & 7.5 & 2.5 & 2.3 & 35.8 & 5.3 & 14.7 & 1.5 & 3.6 & 20.4 & 1.8 & 10.6 & 2.2 & 9.6 & 5.5 & 23.8 & 17.5 & 3.9 \\
\ours & 43.7 & 26.0 & 48.1 & 71.3 & 59.6 & 38.8 & 91.2 & 24.3 & 94.1 & 12.2 & 51.9 & 2.6 & 23.6 & 37.8 & 45.1 & 40.0 & 71.0 \\
\ours w/o Def. & 40.2 & 25.2 & 48.5 & 70.2 & 14.6 & 35.2 & 91.2 & 24.3 & 94.4 & 14.5 & 43.7 & 2.6 & 23.6 & 37.2 & 38.8 & 35.9 & 70.1 \\
\hline
\multicolumn{18}{c}{\cellcolor{gray!20}\textbf{3D Amodal}} \\
\hline
GenPose++ (GP++) & 32.9 & 17.1 & 35.2 & 64.4 & 9.4 & 27.3 & 69.5 & 21.9 & 88.2 & 6.8 & 32.1 & 2.9 & 19.3 & 27.3 & 22.0 & 26.5 & 56.7 \\
MagicPony+GP++ & 7.1 & 0.7 & 0.6 & 27.3 & 4.3 & 9.1 & 58.1 & 0.4 & 17.1 & 1.9 & 17.6 & 3.0 & 4.9 & 5.9 & 10.8 & 11.8 & 4.2 \\
\ours & 40.8 & 22.8 & 41.1 & 74.2 & 35.1 & 39.4 & 79.4 & 27.2 & 89.5 & 5.4 & 45.5 & 5.8 & 21.1 & 33.0 & 46.7 & 39.6 & 67.8 \\
\ours w/o Def. & 37.8 & 21.6 & 41.1 & 71.5 & 8.4 & 34.9 & 79.0 & 26.8 & 89.9 & 6.8 & 42.2 & 3.9 & 21.1 & 34.4 & 23.0 & 39.7 & 67.2 \\
\bottomrule
\end{tabular}
}
\resizebox{\linewidth}{!}{
\begin{tabular}{lllllllllllllllllll}
\toprule
& mean 
& \makecell{hair\\dryer}
& \makecell{ham-\\burger}
& \makecell{hand\\cream}
& handbag
& knife
& lemon
& light
& \makecell{lotus\\root}
& mango
& \makecell{mango-\\steen}
& \makecell{medicine\\bottle}
& mouse
& mug
& orange
& pillow
& \makecell{pome-\\granate}
& \makecell{power\\strip}
\\
\hline
\multicolumn{19}{c}{\cellcolor{gray!20}\textbf{3D}} \\
\hline
GenPose++ (GP++) & 34.3 & 19.6 & 74.8 & 44.0 & 18.0 & 28.0 & 23.3 & 36.5 & 52.7 & 21.5 & 42.6 & 35.9 & 51.4 & 20.2 & 40.8 & 41.5 & 38.1 & 31.6 \\
MagicPony+GP++ & 7.1 & 0.3 & 50.5 & 2.7 & 1.2 & 5.0 & 0.6 & 8.4 & 9.3 & 0.7 & 19.7 & 8.3 & 2.0 & 0.7 & 14.7 & 7.9 & 4.8 & 2.6 \\
\ours & 41.5 & 27.4 & 81.1 & 50.2 & 16.3 & 32.4 & 32.8 & 45.7 & 53.5 & 30.3 & 38.0 & 56.1 & 59.0 & 24.4 & 54.3 & 51.5 & 36.9 & 41.5 \\
\ours w/o Def. & 38.4 & 19.5 & 82.3 & 48.8 & 21.1 & 30.4 & 33.3 & 45.7 & 52.8 & 27.6 & 39.6 & 48.4 & 56.1 & 21.5 & 50.6 & 48.5 & 34.1 & 37.3 \\
\hline
\multicolumn{19}{c}{\cellcolor{gray!20}\textbf{3D Modal}} \\
\hline
DINOv2+D & 24.4 & 18.4 & 36.4 & 24.5 & 11.2 & 16.9 & 25.6 & 34.3 & 28.0 & 23.4 & 24.7 & 45.8 & 17.7 & 4.9 & 57.6 & 24.3 & 33.1 & 24.7 \\
$\text{MagicPony}_{\text{2D}}\text{+D}$  & 14.0 & 5.8 & 40.9 & 13.1 & 4.1 & 13.5 & 7.8 & 24.9 & 19.3 & 9.3 & 12.3 & 21.5 & 6.1 & 1.4 & 11.1 & 16.5 & 23.1 & 8.6 \\
NOCS+D & 26.4 & 14.4 & 60.1 & 37.2 & 2.5 & 15.9 & 19.0 & 43.4 & 40.4 & 17.6 & 18.2 & 50.7 & 48.8 & 12.0 & 27.4 & 42.7 & 9.2 & 28.4 \\
\doublehline
GenPose++ (GP++) & 37.0 & 21.0 & 83.2 & 45.6 & 14.9 & 32.6 & 24.4 & 42.4 & 57.0 & 19.6 & 38.3 & 40.2 & 59.1 & 30.4 & 34.6 & 53.5 & 37.3  & 36.7 \\
MagicPony+GP++ & 7.5 & 0.2 & 54.1 & 3.2 & 1.6 & 6.8 & 1.1 & 12.7 & 14.0 & 1.8 & 27.3 & 4.3 & 3.0 & 0.8 & 5.0 & 8.9 & 6.3 & 3.4 \\
\ours & 43.7 & 29.1 & 81.4 & 53.4 & 14.0 & 33.4 & 36.7 & 62.3 & 57.0 & 34.6 & 33.3 & 63.6 & 64.8 & 33.1 & 41.5 & 59.7 & 26.3& 46.6 \\
\ours w/o Def. & 40.2 & 23.4 & 83.2 & 51.3 & 16.1 & 31.8 & 37.8 & 55.1 & 56.4 & 29.9 & 32.7 & 55.1 & 61.3 & 30.1 & 42.4 & 55.0 & 24.6 &  41.8 \\
\hline
\multicolumn{19}{c}{\cellcolor{gray!20}\textbf{3D Amodal}} \\
\hline
GenPose++ (GP++) & 32.9 & 19.4 & 66.4 & 43.4 & 19.3 & 23.3 & 22.2 & 34.6 & 49.8 & 22.6 & 45.8 & 33.7 & 42.1 & 16.6 & 45.6 & 37.3 & 38.8 & 29.4 \\
MagicPony+GP++ & 7.1 & 0.4 & 46.8 & 2.6 & 1.0 & 3.1 & 0.0 & 7.1 & 6.1 & 0.0 & 14.4 & 10.0 & 0.8 & 0.7 & 22.1 & 7.6 & 3.2 & 2.3 \\
\ours & 40.8 & 27.1 & 80.9 & 49.1 & 17.3 & 31.3 & 28.9 & 40.4 & 51.2 & 27.9 & 41.6 & 52.2 & 52.0 & 21.4 & 64.3 & 48.7 & 46.3 & 39.4 \\
\ours w/o Def. & 37.8 & 18.8 & 81.4 & 47.9 & 23.1 & 29.0 & 28.9 & 42.7 & 50.4 & 26.3 & 44.9 & 44.9 & 49.8 & 18.5 & 57.0 & 46.3 & 42.5 & 35.3 \\
\bottomrule
\end{tabular}
}

\resizebox{\linewidth}{!}{
\begin{tabular}{llllllllllllllllllll}
\toprule
& mean
& remote
& sausage
& shampoo
& shoe
& shrimp
& teapot
& \makecell{tooth\\brush}
& \makecell{tooth\\paste}
& \makecell{toy\\animal}
& \makecell{toy\\boat}
& \makecell{toy\\bus}
& \makecell{toy\\car}
& \makecell{toy\\m'bike}
& \makecell{toy\\plane}
& \makecell{toy\\train}
& \makecell{toy\\truck}
& wallet
\\
\hline
\multicolumn{19}{c}{\cellcolor{gray!20}\textbf{3D}} \\
\hline
GenPose++ (GP++) & 34.3 & 40.9 & 24.1 & 90.1 & 62.9 & 12.3 & 16.0 & 67.2 & 65.2 & 1.6 & 14.4 & 37.1 & 2.2 & 17.9 & 18.2 & 23.6 & 30.7 & 24.9 \\
MagicPony+GP++ & 7.1 & 1.7 & 1.9 & 19.5 & 7.2 & 1.4 & 1.0 & 0.4 & 4.0 & 2.2 & 0.9 & 0.9 & 0.7 & 2.2 & 2.1 & 1.5 & 3.2 & 2.8 \\
\ours & 41.5 & 44.7 & 31.0 & 92.8 & 62.9 & 16.4 & 26.4 & 72.8 & 69.4 & 5.4 & 21.0 & 47.9 & 3.3 & 22.5 & 26.0 & 31.5 & 37.3 & 31.1 \\
\ours w/o Def. & 38.4 & 44.2 & 31.9 & 91.4 & 57.6 & 15.5 & 15.3 & 73.0 & 68.5 & 1.6 & 16.5 & 41.3 & 1.9 & 19.8 & 21.6 & 30.2 & 33.5 & 30.5 \\
\hline
\multicolumn{19}{c}{\cellcolor{gray!20}\textbf{3D Modal}} \\
\hline
DINOv2+D & 24.4 & 15.4 & 30.6 & 52.1 & 30.0 & 14.0 & 11.4 & 51.0 & 41.4 & 14.8 & 7.5 & 14.4 & 11.6 & 11.0 & 9.2 & 11.6 & 20.2 & 20.6 \\
$\text{MagicPony}_{\text{2D}}\text{+D}$  & 14.0 & 12.0 & 22.2 & 31.5 & 11.5 & 7.4 & 6.6 & 12.9 & 29.2 & 4.9 & 2.7 & 4.7 & 2.7 & 10.1 & 3.1 & 5.0 & 8.8 & 6.9 \\
NOCS+D & 26.4 & 28.7 & 13.0 & 71.4 & 39.0 & 5.5 & 12.0 & 66.8 & 60.6 & 1.8 & 4.5 & 34.6 & 0.3 & 7.4 & 13.5 & 25.2 & 14.7 & 16.8 \\
\doublehline
GenPose++ (GP++) & 37.0 & 45.0 & 28.7 & 92.5 & 64.6 & 11.7 & 21.5 & 80.7 & 67.4 & 2.5 & 12.9 & 38.5 & 2.4 & 27.5 & 14.9 & 32.9 & 22.8 & 25.2 \\
MagicPony+GP++ & 7.5 & 2.0 & 0.9 & 21.9 & 11.9 & 0.8 & 1.0 & 0.4 & 3.0 & 1.5 & 1.1 & 0.3 & 1.2 & 4.1 & 2.1 & 1.1 & 3.4 & 4.9 \\
\ours & 43.7 & 49.7 & 33.3 & 87.0 & 58.5 & 17.5 & 31.5 & 82.3 & 72.9 & 3.7 & 19.9 & 49.2 & 3.1 & 33.8 & 23.6 & 41.2 & 29.8 & 30.5 \\
\ours w/o Def. & 40.2 & 48.2 & 33.3 & 87.7 & 50.8 & 14.4 & 16.8 & 82.8 & 71.9 & 1.2 & 17.2 & 44.5 & 1.0 & 27.8 & 16.8 & 40.9 & 25.4 & 30.5 \\
\hline
\multicolumn{19}{c}{\cellcolor{gray!20}\textbf{3D Amodal}} \\
\hline
GenPose++ (GP++) & 32.9 &  39.2 & 19.4 & 87.7 & 61.9 & 12.5 & 13.7 & 57.9 & 64.4 & 1.4 & 14.8 & 36.7 & 2.1 & 15.0 & 19.2 & 20.1 & 34.0 & 24.8 \\
MagicPony+GP++ & 7.1 & 1.6 & 2.8 & 17.1 & 3.8 & 1.7 & 1.0 & 0.5 & 4.4 & 2.4 & 0.9 & 1.1 & 0.6 & 1.6 & 2.1 & 1.6 & 3.1 & 1.6 \\
\ours & 40.8 &  42.6 & 28.7 & 98.6 & 65.7 & 16.0 & 24.3 & 66.3 & 68.1 & 5.9 & 21.3 & 47.5 & 3.3 & 19.0 & 26.7 & 27.8 & 40.5 & 31.4 \\
\ours w/o Def. & 37.8 & 42.5 & 30.6 & 95.2 & 61.9 & 15.9 & 14.7 & 66.3 & 67.3 & 1.7 & 16.3 & 40.3 & 2.1 & 17.3 & 23.1 & 26.2 & 36.9 & 30.6 \\
\bottomrule
\end{tabular}

}
\label{supptab:all_results_3D}
\end{table}

\begin{table}[t]
\footnotesize
\centering
\caption{\textbf{PCK@0.1 results} for 2D category-level correspondences comparison across all 50 classes for \ours and the baselines. Beyond showcasing which categories are almost solved versus still challenging, the table reveals how object variability drives performance: classes with low deformation and consistent shapes (\eg, \emph{shampoo}, \emph{corn}) are nearly saturated, whereas highly diverse toy categories (\eg, \emph{toy car}, \emph{toy animal}) remain difficult.}
\vspace{-1em}
\resizebox{\linewidth}{!}{
\begin{tabular}{llllllllllllllllll}
\toprule
 & mean & backpack & book & bottle & box & bread & coconut & conch & corn & dinosaur & dish & doll & egg & eraser & \makecell{facial\\cream} & \makecell{flower\\pot} & \makecell{glasses\\case} \\
\hline
\multicolumn{18}{c}{\cellcolor{gray!20}\textbf{2D}} \\
\hline
DINOv2 & 22.9 & 7.0 & 14.9 & 25.9 & 41.1 & 14.0 & 30.9 & 15.8 & 30.8 & 16.4 & 8.6 & 5.0 & 12.7 & 64.4 & 11.7 & 9.5 & 44.2 \\
$\text{MagicPony}_{\text{2D}}$  & 15.7 & 6.4 & 7.1 & 36.9 & 41.8 & 22.9 & 22.1 & 5.0 & 21.8 & 7.4 & 10.2 & 5.6 & 11.2 & 36.9 & 14.4 & 9.7 & 38.1 \\
NOCS & 26.7 & 27.2 & 28.4 & 35.8 & 49.7 & 23.7 & 59.3 & 2.4 & 57.0 & 8.0 & 3.0 & 6.7 & 1.5 & 35.1 & 12.1 & 4.5 & 55.2 \\
\doublehline
GenPose++ (GP++) & 36.3 & 37.0 & 43.2 & 42.8 & 31.1 & 30.0 & 70.4 & 12.2 & 73.6 & 13.3 & 13.2 & 4.3 & 11.6 & 40.8 & 27.3 & 16.0 & 64.3 \\
\hline 
MagicPony+GP++ & 10.7 & 4.8 & 4.5 & 20.2 & 21.8 & 22.1 & 32.0 & 1.9 & 16.8 & 8.1 & 9.1 & 6.9 & 8.2 & 21.2 & 12.0 & 8.1 & 13.6 \\
\ours & 41.2 & 40.9 & 46.2 & 47.2 & 61.8 & 36.5 & 73.3 & 16.0 & 75.2 & 15.1 & 15.6 & 8.5 & 15.4 & 46.5 & 26.7 & 18.9 & 73.1 \\
\ours w/o Def. & 39.1 & 39.9 & 46.9 & 46.3 & 34.6 & 34.2 & 73.3 & 13.2 & 74.7 & 14.8 & 15.2 & 5.0 & 16.1 & 47.4 & 26.6 & 19.1 & 72.5 \\
\bottomrule
\end{tabular}
}
\resizebox{\linewidth}{!}{
\begin{tabular}{lllllllllllllllllll}
\toprule
& mean 
& \makecell{hair\\dryer}
& \makecell{ham-\\burger}
& \makecell{hand\\cream}
& handbag
& knife
& lemon
& light
& \makecell{lotus\\root}
& mango
& \makecell{mango-\\steen}
& \makecell{medicine\\bottle}
& mouse
& mug
& orange
& pillow
& \makecell{pome-\\granate}
& \makecell{power\\strip}
\\
\hline
\multicolumn{19}{c}{\cellcolor{gray!20}\textbf{2D}} \\
\hline
DINOv2 & 22.9 & 19.2 & 18.0 & 30.0 & 16.9 & 15.6 & 31.7 & 16.0 & 28.5 & 23.9 & 26.3 & 17.6 & 17.9 & 13.2 & 33.3 & 39.0 & 21.8 & 22.7 \\
$\text{MagicPony}_{\text{2D}}$  & 15.7 & 6.8 & 36.8 & 19.7 & 8.2 & 13.0 & 11.1 & 17.4 & 21.1 & 11.8 & 12.8 & 13.5 & 6.4 & 7.0 & 11.1 & 22.2 & 12.7 & 13.1 \\
NOCS & 26.7 & 21.9 & 67.3 & 43.2 & 17.7 & 11.6 & 20.3 & 19.3 & 28.0 & 28.0 & 3.0 & 24.9 & 53.4 & 26.5 & 24.6 & 47.7 & 9.3 & 27.8 \\
\doublehline
GenPose++ (GP++) & 36.3 & 27.0 & 76.4 & 48.5 & 34.0 & 26.8 & 29.4 & 37.0 & 39.9 & 35.4 & 27.9 & 31.7 & 58.9 & 37.0 & 30.6 & 59.3 & 30.2 & 37.0 \\
\hline
MagicPony+GP++ & 10.7 & 4.7 & 55.7 & 11.2 & 7.5 & 9.0 & 3.3 & 12.1 & 8.6 & 5.1 & 15.4 & 9.6 & 5.2 & 5.1 & 15.8 & 22.7 & 7.5  & 8.6 \\
\ours & 41.2 & 35.1 & 79.3 & 55.8 & 31.0 & 30.7 & 33.9 & 37.6 & 39.7 & 43.4 & 21.8 & 39.1 & 63.6 & 43.3 & 38.9 & 65.2 & 26.6 & 44.5 \\
\ours w/o Def. & 39.1 & 28.8 & 80.5 & 54.3 & 36.7 & 29.3 & 34.4 & 40.5 & 40.3 & 43.8 & 21.0 & 36.2 & 61.2 & 39.6 & 35.5 & 63.8 & 24.2 & 41.2 \\
\bottomrule
\end{tabular}
}

\resizebox{\linewidth}{!}{
\begin{tabular}{llllllllllllllllllll}
\toprule
& mean
& remote
& sausage
& shampoo
& shoe
& shrimp
& teapot
& \makecell{tooth\\brush}
& \makecell{tooth\\paste}
& \makecell{toy\\animal}
& \makecell{toy\\boat}
& \makecell{toy\\bus}
& \makecell{toy\\car}
& \makecell{toy\\m'bike}
& \makecell{toy\\plane}
& \makecell{toy\\train}
& \makecell{toy\\truck}
& wallet
\\
\hline
\multicolumn{19}{c}{\cellcolor{gray!20}\textbf{2D}} \\
\hline
DINOv2 & 22.9 & 19.1 & 35.2 & 56.5 & 30.0 & 23.0 & 13.1 & 37.7 & 41.1 & 14.9 & 17.1 & 13.3 & 9.6 & 10.6 & 15.2 & 14.5 & 13.0 & 25.4 \\
$\text{MagicPony}_{\text{2D}}$  & 15.7 & 15.2 & 20.4 & 49.0 & 13.8 & 9.5 & 8.9 & 8.7 & 29.6 & 8.1 & 8.8 & 7.2 & 5.1 & 9.1 & 7.7 & 11.7 & 8.5 & 12.5 \\
NOCS & 26.7 & 27.7 & 15.0 & 69.7 & 45.2 & 9.5 & 21.7 & 52.2 & 55.9 & 0.9 & 14.0 & 42.6 & 5.5 & 16.3 & 20.7 & 27.3 & 28.5 & 28.2 \\
\doublehline
GenPose++ (GP++) & 36.3 & 41.2 & 25.0 & 90.8 & 62.6 & 17.5 & 23.1 & 62.2 & 65.4 & 6.5 & 20.5 & 50.2 & 7.1 & 26.7 & 28.8 & 34.4 & 41.5 & 35.8 \\
\hline 
MagicPony+GP++ & 10.7 & 4.5 & 2.3 & 22.6 & 11.9 & 4.3 & 3.9 & 0.4 & 11.9 & 4.7 & 4.1 & 4.2 & 3.8 & 8.1 & 7.2 & 5.7 & 7.7 & 11.5 \\
\ours & 41.2 & 45.6 & 31.0 & 89.7 & 62.9 & 21.4 & 37.3 & 68.3 & 68.9 & 8.9 & 28.1 & 57.1 & 9.2 & 31.3 & 34.8 & 43.3 & 47.1 & 39.6 \\
\ours w/o Def.& 39.1  & 44.9 & 32.4 & 91.8 & 59.1 & 18.2 & 25.0 & 68.3 & 68.5 & 7.9 & 22.5 & 51.8 & 6.7 & 29.5 & 32.0 & 41.8 & 44.9 & 38.1 \\
\bottomrule
\end{tabular}

}
\label{supptab:all_results_2D}
\end{table}

\begin{table}[ht]
  \centering
    \caption{Maximum number of annotated keypoints observed for each category.}\label{supptab:nb_keypoints}
    \setlength{\tabcolsep}{12pt}
  \begin{tabular}{lrlr}
    \toprule
    Category & \#Keypoints & Category & \#Keypoints \\
    \midrule
    backpack & 16 & medicine\_bottle & 6 \\
    book & 8 & mouse & 7 \\
    bottle & 5 & mug & 14 \\
    box & 14 & orange & 3 \\
    bread & 6 & pillow & 6 \\
    coconut & 2 & pomegranate & 4 \\
    conch & 7 & power\_strip & 10 \\
    corn & 3 & remote\_control & 11 \\
    dinosaur & 18 & sausage & 2 \\
    dish & 9 & shampoo & 2 \\
    doll & 11 & shoe & 12 \\
    egg & 3 & shrimp & 8 \\
    eraser & 10 & teapot & 13 \\
    facial\_cream & 12 & tooth\_brush & 8 \\
    flower\_pot & 10 & tooth\_paste & 7 \\
    glasses\_case & 16 & toy\_animals & 11 \\
    hair\_dryer & 14 & toy\_boat & 8 \\
    hamburger & 2 & toy\_bus & 18 \\
    hand\_cream & 7 & toy\_car & 8 \\
    handbag & 15 & toy\_motorcycle & 19 \\
    knife & 6 & toy\_plane & 13 \\
    lemon & 2 & toy\_train & 10 \\
    light & 18 & toy\_truck & 10 \\
    lotus\_root & 3 & wallet & 9 \\
    mango & 3 & mangosteen & 4 \\
    \bottomrule
  \end{tabular}
\end{table}

\section*{Reproducibility and LLM assistance}
The complete processing pipeline, including scripts for dataset preparation and annotation generation, is available at \href{https://github.com/GenIntel/HouseCorr3D}{\faGithub/GenIntel/HouseCorr3D}. The dataset, including the annotated 3D meshes and projected 2D keypoints, is accessible from \href{http://huggingface.co/}{Hugging Face} for easy access and long-term hosting. In addition, we provide helper functions to compute the 3D correspondence metrics introduced in this paper, ensuring that results can be evaluated in a consistent and standardized manner.

We used large language models (LLMs) in a limited capacity to assist with the writing of this paper. Specifically, LLMs were employed only to (i) improve sentence clarity and conciseness, and (ii) condense overly lengthy paragraphs. All technical contributions — including the method design, experimental setup, results, and analyses — are entirely our own work.

\begin{figure}[p]
    \centering
    \includegraphics[width=\textwidth, keepaspectratio]{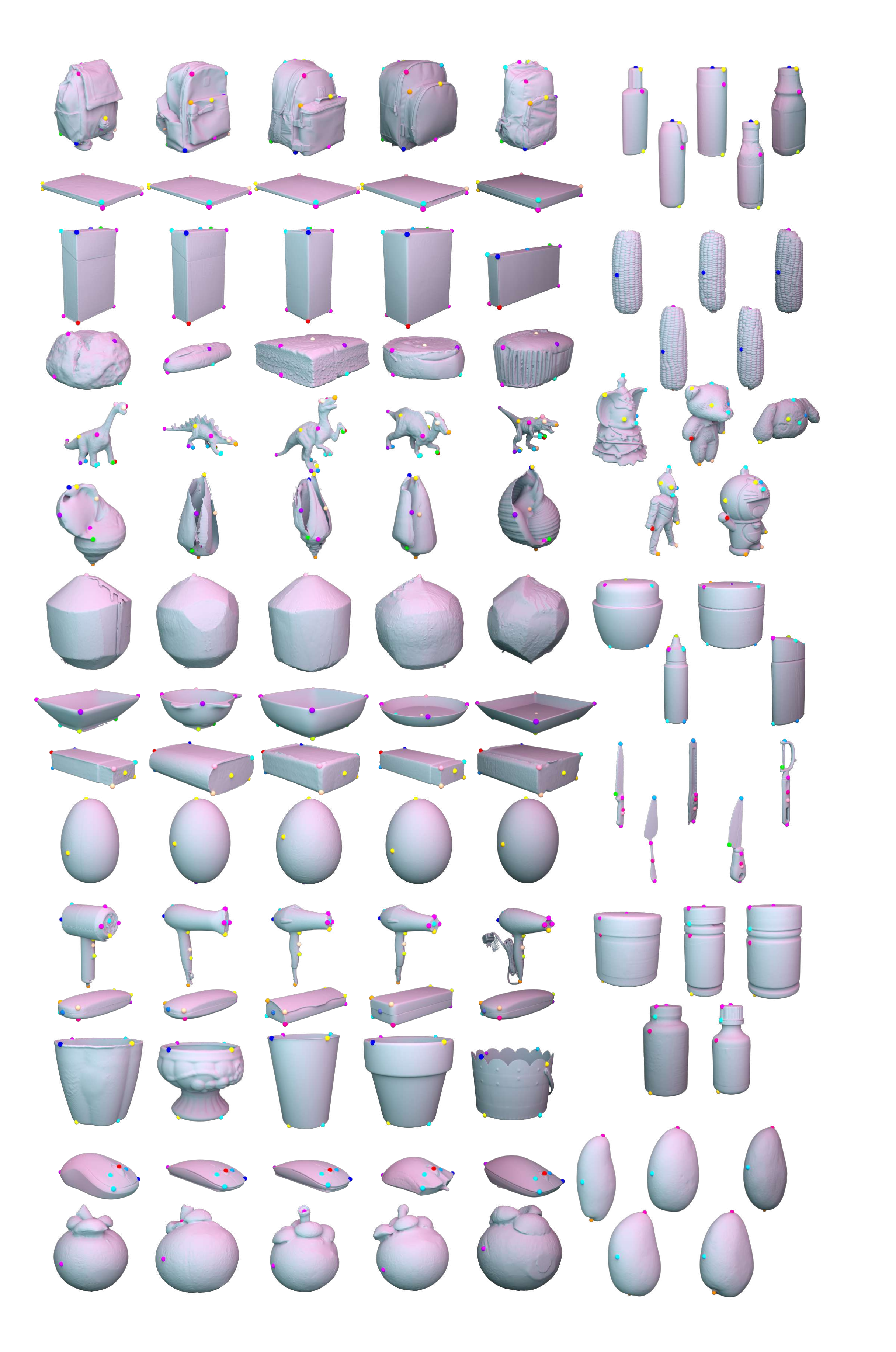}
    \caption{Full keypoints annotation overview of \ourDataset (part 1 of 3). \\Note than some keypoints color variation can be due to lighting effects.}
    \label{suppfig:full_data_overview}
\end{figure}

\begin{figure}[p]\ContinuedFloat
    \centering
    \includegraphics[width=\textwidth, keepaspectratio]{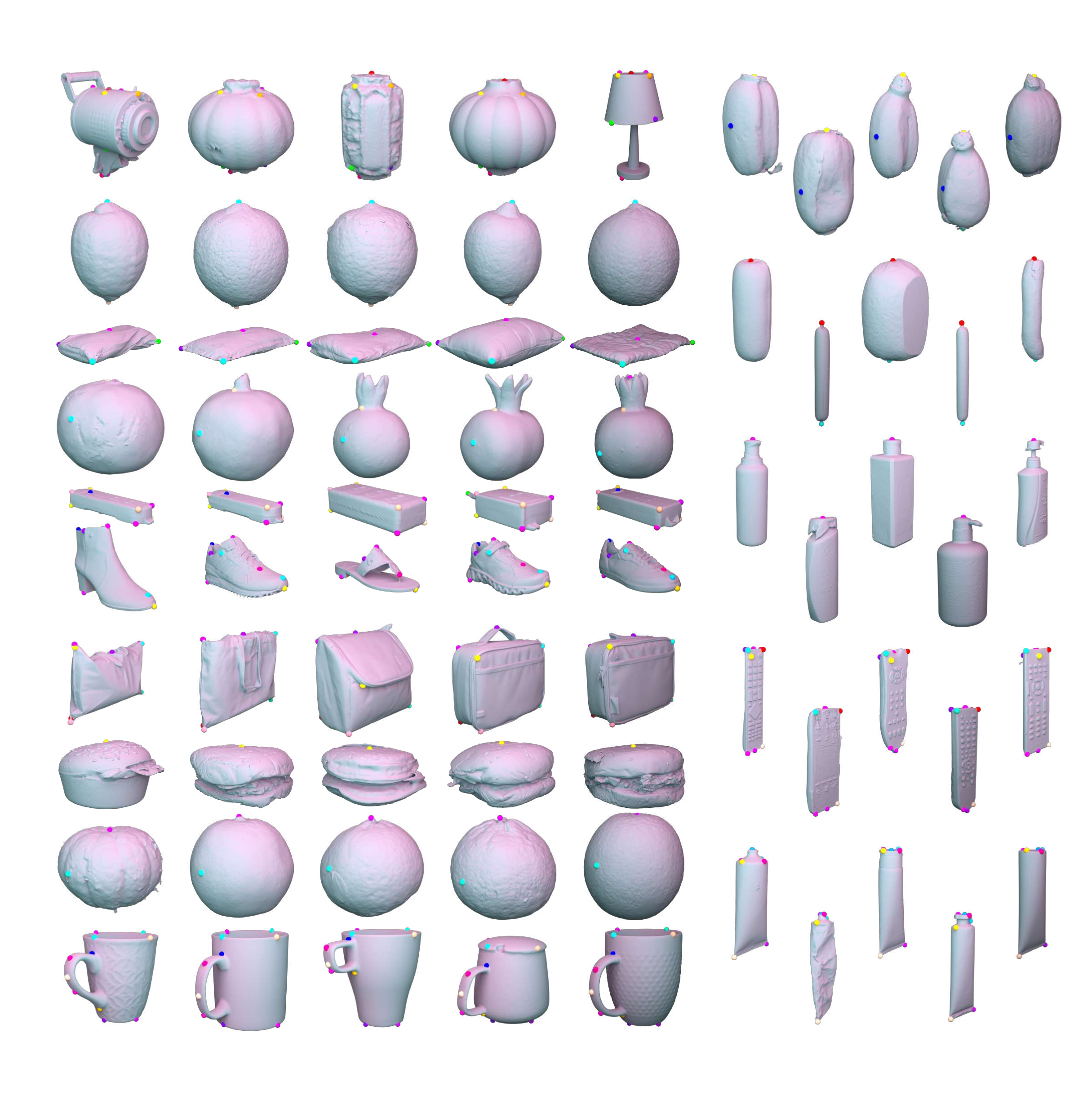}
    \caption{Full keypoints annotation overview of \ourDataset (part 2 of 3). \\Note than some keypoints color variation can be due to lighting effects.}
\end{figure}

\begin{figure}[p]\ContinuedFloat
    \centering
    \includegraphics[width=\textwidth, keepaspectratio]{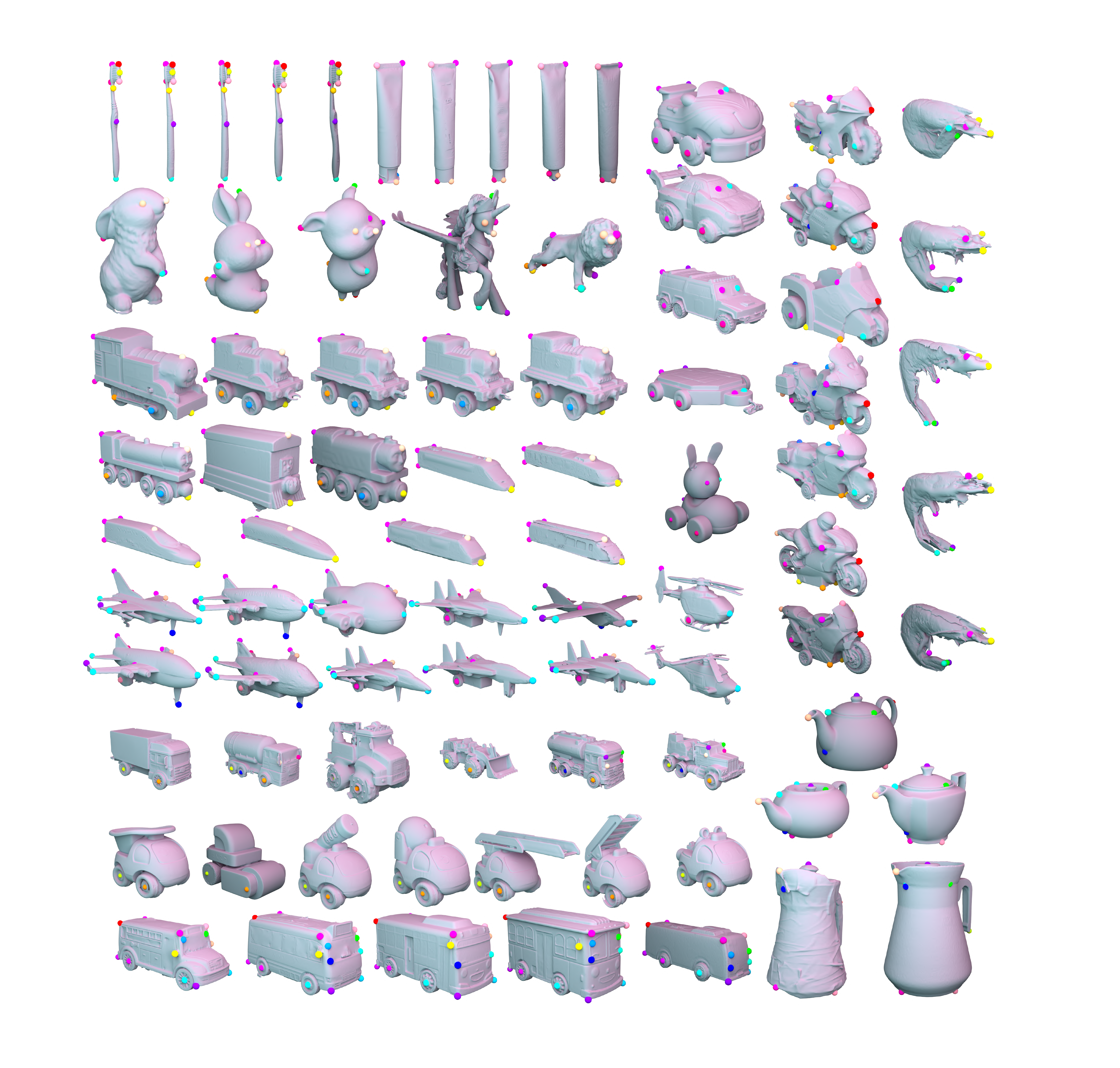}
    \caption{Full keypoints annotation overview of \ourDataset (part 3 of 3). \\Note than some keypoints color variation can be due to lighting effects.}
\end{figure}

\end{document}